\documentclass{article}%
\usepackage{amsmath}
\usepackage{amsfonts}
\usepackage{amssymb}
\usepackage{graphicx}
\providecommand{\U}[1]{\protect \rule{.1in}{.1in}}

\providecommand{\U}[1]{\protect \rule{.1in}{.1in}}

\begin{document}

\title{SurvBeNIM: The Beran-Based Neural Importance Model for Explaining the Survival Models}
\author{Lev V. Utkin, Danila Y. Eremenko and Andrei V. Konstantinov\\Higher School of Artificial Intelligence Technologies \\Peter the Great St.Petersburg Polytechnic University\\St.Petersburg, Russia\\e-mail: lev.utkin@gmail.com, danilaeremenko@mail.ru, andrue.konst@gmail.com}
\date{}
\maketitle

\begin{abstract}
A new method called the Survival Beran-based Neural Importance Model
(SurvBeNIM) is proposed. It aims to explain predictions of machine learning
survival models, which are in the form of survival or cumulative hazard
functions. The main idea behind SurvBeNIM is to extend the Beran estimator by
incorporating the importance functions into its kernels and by implementing
these importance functions as a set of neural networks which are jointly
trained in an end-to-end manner. Two strategies of using and training the
whole neural network implementing SurvBeNIM are proposed. The first one
explains a single instance, and the neural network is trained for each
explained instance. According to the second strategy, the neural network only
learns once on all instances from the dataset and on all generated instances.
Then the neural network is used to explain any instance in a dataset domain.
Various numerical experiments compare the method with different existing
explanation methods. A code implementing the proposed method is publicly available.


\end{abstract}

\section{Introduction}

One of the important types of data in several applications is censored
survival data processed in the framework of survival analysis
\cite{Hosmer-Lemeshow-May-2008,Wang-Li-Reddy-2019}. This type of data can be
found in applications where objects are characterized by times to some events
of interest, for example, by times to failure in reliability, times to
recovery or times to death in medicine, times to bankruptcy of a bank or times
to an economic crisis in economics. The important peculiarity of survival data
is that the corresponding event does not necessarily occur during its
observation period. In this case, we say about the so-called censored or
right-censored data \cite{Nezhad-etal-2018}.

There are many machine learning models dealing with survival data, including
models based on applying and extending the Cox proportional hazard model
\cite{Cox-1972}, for example, models presented in
\cite{Witten-Tibshirani-2010,Zhu-Yao-Huang-2016}, models based on a survival
modification of random forests and called random survival forests (RSF)
\cite{Ibrahim-etal-2008,Mogensen-etal-2012,Schmid-etal-2016,Wang-Zhou-2017,Wright-etal-2017}%
, models extending the neural networks
\cite{Zhu-Yao-Huang-2016,Haarburger-etal-2018,Katzman-etal-2018,Wiegrebe-etal-23}%
.

These models have gained considerable attention for their ability to analyze
time-to-event data and to predict survival outcomes accurately. However, most
models are perceived as black boxes, lacking interpretability. Understanding
the factors influencing the survival predictions is crucial in many
applications, especially, in medicine where a doctor using an intelligence
diagnostic system has to understand a diagnosis stated by the system in order
to be convinced of the diagnosis \cite{Holzinger-etal-2019}. Therefore, there
is a need for the reliable explanation methods that could answer the question
of what features or factors influenced the predictions of the corresponding
machine learning model.

Many explanation methods have been developed and proposed to answer the above
question
\cite{Adadi-Berrada-2018,Arrieta-etal-2020,Bodria-etal-23,Burkart-Huber-21,Carvalho-etal-2019,Cwiling-etal-23,Guidotti-2019,Guidotti-21,Molnar-2019,Murdoch-etal-19a,Notovich-etal-23,Rudin-2019}%
. We point out two the well-known methods and their extensions. The first one
is the Local Interpretable Model-agnostic Explanations (LIME)
\cite{Ribeiro-etal-2016}, which is based on the local linear approximation of
a black-box prediction function. The approximation is carried out by
generating a set of neighborhood instances around the explained instance and
computing the corresponding predictions produced by the black-box model.
Coefficients of the linear approximated function trained on the generated
instances and predictions can be regarded as quantitative measures of the
feature importances. The second important method is the SHapley Additive
exPlanations (SHAP) \cite{Lundberg-Lee-2017,Strumbel-Kononenko-2010}, which
uses a game-theoretic approach and Shapley values regarded as measures of the
feature importances.

However, these methods as well as many other explanation methods mainly deal
with models whose outputs are point-valued. In contrast to these models,
survival machine learning models often predict functions, for example, the
survival function (SF) or the cumulative hazard function (CHF). Therefore, the
extensions of LIME and SHAP, called SurvLIME and SurvSHAP, respectively, have
been proposed to explain the survival black-box models
\cite{Kovalev-Utkin-Kasimov-20a,Utkin-Kovalev-Kasimov-20c,Krzyzinski-etal-23,Utkin-Satyukov-Konstantinov-22}%
. In particular, SurvLIME aims to approximate the black-box model at an
explained point by the Cox model which is based on the linear combination of
features. Coefficients of the linear combination are regarded as measures of
the feature impact on the prediction in the form of the SF or the CHF. Another
interesting method, SurvSHAP \cite{Krzyzinski-etal-23}, is based on adapting
the SHAP method to the functional output of a survival black-box model. In
contrast to SurvLIME, SurvSHAP implements the time-dependent explainability,
i.e. it explains the black-model prediction (the SF) at each time point.

In addition to LIME, we have to point another explanation method called the
Neural Additive Model (NAM) \cite{Agarwal-etal-21}. It can be regarded as an
extension of LIME to some extend when the linear approximation is replaced
with the neural network implementation of the generalized additive model (GAM)
\cite{Hastie-Tibshirani-1990}. The so-called \emph{shape functions} of GAM
produced by NAM show how a model prediction depends on each feature. A
remarkable feature of NAM is that functions in GAM are inferred by training
the corresponding neural network. Following this model, its survival
modification, called SurvNAM, has been proposed in
\cite{Utkin-Satyukov-Konstantinov-22} where the loss function for training the
network minimizes the distance between SFs produced by the black-box model and
by the extended Cox model with GAM instead of the linear combination of
features. SurvNAM can be regarded as an extension of SurvLIME as well as NAM,
namely, it extends SurvLIME by considering the GAM model instead of the linear
combination of features, and it extends NAM by considering the survival models.

In order to overcome some problems related to the possible violation of the
Cox model assumptions (the proportional hazards assumption and the linear
relationship of features), Utkin et al. \cite{Utkin-Eremenko-Konstantinov-23}
proposed to apply a modification of the Beran estimator \cite{Beran-81}
instead of the Cox model in SurvLIME. The Beran estimator is a kernel
extension of the Kaplan-Meier model taking into account the training data
structure. The proposed method, called the Survival Beran eXplanation
(SurvBeX), has demonstrated outperforming results in comparison with SurvLIME
and SurvSHAP.

Although SurvBeX overcomes assumptions of the Cox model (the proportional
hazards assumption and the linear relationship between features) used in
SurvLIME, it computes the linear regression coefficients like SurvLIME, which
measure the feature impacts on predictions. Therefore, we propose a new method
called the \textbf{Surv}ival \textbf{Be}ran-based \textbf{N}eural
\textbf{I}mportance \textbf{M}odel (SurvBeNIM) which combines the advantages
of SurvNAM and SurvBeX. The main idea behind SurvBeNIM is to extend the Beran
estimator by incorporating the so-called \emph{importance functions} into its
kernels and by implementing these importance functions in the form of a set of
neural networks to make them flexible and universal. Similarly to SurvNAM, the
proposed method is based on applying a set of neural networks implementing
functions characterizing the feature impacts on predictions (shape functions
in SurvNAM and the importance functions SurvBeNIM). However, SurvBeNIM does
not use NAM and GAM. In contrast to SurvNAM, the proposed SurvBeNIM does not
consider the shape function. SurvBeNIM aims to find the importance functions
of every feature, which can be viewed as functions whose values show how the
corresponding features impact on a prediction of the black-box model. The
importance functions play a role of functions indicating the feature
importances, but they are not additive. Therefore, we cannot use the name NAM
in the proposed method and use the name \textquotedblleft Neural Importance
Model\textquotedblright \ (NIM).

Our contributions can be summarized as follows:

\begin{enumerate}
\item A new method SurvBeNIM for explaining predictions of machine learning
survival models is proposed, which simultaneously extends and combines SurvNAM
and SurvBeX. According to SurvBeNIM, the importance functions are incorporated
into kernels of the Beran estimator and are implemented as a set of neural
subnetworks which are jointly trained in an end-to-end manner.

\item Two strategies of using and training the neural network implementing
SurvBeNIM are proposed. The first one requires to train the neural network for
each explained instance. The second strategy can be regarded as a global
explanation to some extent. According to this strategy, the neural network
only learns once and then can be used for any instances.

\item A method for improving the training process of the neural network
implementing SurvBeNIM is introduced, which overcomes difficulties of
computing the distance between SFs in areas of small and large times, i.e.
when SFs are close to 1 or to 0.

\item Various numerical experiments compare the proposed method with SurvNAM
as well as SurvLIME under different conditions. The corresponding code
implementing the proposed method is publicly available at: https://github.com/DanilaEremenko/SurvBeNIM.
\end{enumerate}

Numerical results show that SurvBeNIM provides outperforming results in
comparison with other methods.

The paper is organized as follows. Related work considering the existing
explanation methods can be found in Section 2. A short description of basic
concepts of survival analysis, including the Cox model and the Beran
estimator, is given in Section 3. Explanation methods such as LIME and NAM are
described in Section 4. A short description of the explanation methods for
survival models, including SurvLIME, SurvNAM, SurvBeX, can be found in Section
5. A general idea of SurvBeNIM and its two realizations are provided in
Section 6. Numerical experiments with synthetic and real data are given in
Section 7. Concluding remarks are provided in Section 8.

\section{Related work}

\textbf{Explanation methods.} The need to explain the prediction results of
many machine learning models, especially, in the critical applications such as
medicine, security, etc., has led to the fact that a huge number of
explanation methods have been developed. Various surveys are devoted to study
the explanation methods and their comparison
\cite{Adadi-Berrada-2018,Arrieta-etal-2020,Bodria-etal-23,Burkart-Huber-21,Carvalho-etal-2019,Cwiling-etal-23,Guidotti-2019,Molnar-2019,Murdoch-etal-19a,Notovich-etal-23,Rudin-2019}%
. Let us consider three important groups of explanation methods which are used
for developing and comparing the proposed SurvBeNIM method.

The first group consists of modifications of LIME \cite{Ribeiro-etal-2016} and
includes the following methods: Anchor LIME \cite{Ribeiro-etal-2018}, ALIME
\cite{Shankaranarayana-Runje-2019}, DLIME \cite{Zafar-Khan-2019}, GraphLIME
\cite{Huang-Yamada-etal-2020}, LIME-Aleph \cite{Rabold-etal-2019}, LIME-SUP
\cite{Hu-Chen-Nair-Sudjianto-2018}, NormLIME \cite{Ahern-etal-2019}, Rank-LIME
\cite{Chowdhury-etal-22}, s-LIME \cite{Gaudel-etal-22}. Important papers
studying the theoretical basis of LIME and its modifications are presented by
Garreau, von Luxburg, Mardaoui
\cite{Garreau-23,Garreau-Luxburg-2020,Garreau-Luxburg-2020a,Garreau-Mardaoui-21}%
.

The second group of explanation methods is formed around SHAP
\cite{Lundberg-Lee-2017,Strumbel-Kononenko-2010}. Similarly to LIME, a lot of
modifications of SHAP have been developed, including K-SHAP
\cite{Coletta-etal-23}, SHAP-E \cite{Jun-Nichol-23}, SHAP-IQ
\cite{Fumagalli-etal-23}, Random SHAP \cite{Utkin-Konstantinov-22n}, Latent
SHAP \cite{Bitto-etal-22}, MM-SHAP \cite{Parcalabescu-Frank-22}. The
aforementioned modifications are a small part of all extensions of SHAP. A
thorough theoretical study of basic concepts of SHAP is presented in
\cite{Broeck-etal-22}.

The third group of explanation methods can be regarded as a generalization of
LIME when the linear regression functions approximating the black-box model
are replaced with GAM \cite{Hastie-Tibshirani-1990} because it is a more
general and flexible model in comparison with the original linear model. In
GAM, the so-called shape functions of features can explain how each feature
contributes into a prediction. The most interesting explanation method
realizing the idea of GAM is NAM \cite{Agarwal-etal-20} which implements shape
functions in the form of neural networks. Various realizations of GAM in
explanation methods have led to a number of interesting methods, including
\cite{Chang-Tan-etal-2020,Chen-Vaughan-etal-20,Konstantinov-Utkin-21,Lou-etal-12,Nori-etal-19,Radenovic-etal-22,Yang-Zhang-Sudjianto-20,Zhang-Tan-Koch-etal-19}%
.

An important statistical generalization of many explanation methods was
proposed in \cite{Senetaire-etal-23}.

\textbf{Explanation methods in survival analysis. }Most survival machine
learning models can be regarded as black boxes inferring predictions in the
form of SFs or CHFs which have to be explained in many applications to enhance
the belief to the models. The form of predictions and peculiarities of
survival models require and motivate to develop the explanation methods which
take into account the predictions and the peculiarities. A method called
SurvLIME has been presented in \cite{Kovalev-Utkin-Kasimov-20a} where an idea
of using the Cox model for explaining the black-box model prediction was
proposed. SurvLIME has been modified in \cite{Utkin-Kovalev-Kasimov-20c} where
the Euclidean distance between SFs was replaced with the Chebyshev Distance.
The idea of using the Cox model was applied to the counterfactual explanation
of machine learning survival models in \cite{Kovalev-Utkin-etal-21}. A method
for the uncertainty interpretation of the survival model predictions was
considered in \cite{Utkin-etal-21e}. A modification of SurvLIME called
SurvLIMEpy and the software package implementing the method were presented in
\cite{Pachon-Garcia-etal-23}.

A modification of NAM \cite{Agarwal-etal-20} for survival models called
SurvNAM has been proposed in \cite{Utkin-Satyukov-Konstantinov-22}. Another
extension of NAM for survival analysis with EHR data was presented in
\cite{Peroni-etal-22}. An approach called EXplainable CEnsored Learning for
explaining models in the framework of survival analysis has been presented in
\cite{Wu-Peng-etal-22}. An application-based machine learning explainable
model using for the breast cancer survival analysis has been proposed in
\cite{Moncada-Torres-etal-21}. The first method based on the SHAP extension
called SurvSHAP(t) for explaining the survival model predictions has been
developed and presented in \cite{Krzyzinski-etal-23}.

\section{Basic concepts of survival analysis}

Instances in survival analysis are represented by a triplet $(\mathbf{x}%
_{i},\delta_{i},T_{i})$, where $\mathbf{x}_{i}^{\mathrm{T}}=(x_{i}%
^{(1)},...,x_{i}^{(d)})\in \mathbb{R}^{d}$ is a vector of $d$ features
characterizing the $i$-th object whose time to an event of interest is $T_{i}%
$. $\delta_{i}$ is the indicator function taking the value 1 if the event of
interest is observed, and 0 if the event is not observed and the end of the
observation is used. When $\delta_{i}=0$, we have a censored observation
\cite{Hosmer-Lemeshow-May-2008}.

Survival analysis aims to estimate the time to the event $T$ for a new
instance $\mathbf{x}$ by using the training set $\mathcal{A}=\{(\mathbf{x}%
_{i},\delta_{i},T_{i}),i=1,...,n\}$.

The SF denoted as $S(t|\mathbf{x})$ is the probability of surviving up to time
$t$, that is $S(t|\mathbf{x})=\Pr \{T>t|\mathbf{x}\}$. The CHF denoted as
$H(t|\mathbf{x})$ is expressed through the SF as follows:
\begin{equation}
S(t|\mathbf{x})=\exp \left(  -H(t|\mathbf{x})\right)  .
\end{equation}

The C-index proposed by Harrell et al. \cite{Harrell-etal-1982} is used for
comparison of different survival models. It estimates the probability that the
event times of a pair of instances are correctly ranking. One of the forms of
the C-index, which will be used below, is \cite{Uno-etal-11}:
\begin{equation}
C=\frac{\sum \nolimits_{i,j}\mathbf{1}[T_{i}<T_{j}]\cdot \mathbf{1}[\widehat
{T}_{i}<\widehat{T}_{j}]\cdot \delta_{i}}{\sum \nolimits_{i,j}\mathbf{1}%
[T_{i}<T_{j}]\cdot \delta_{i}},
\end{equation}
where $\widehat{T}_{i}$ and $\widehat{T}_{j}$ are the predicted survival
durations; $\mathbf{1}[T_{i}<T_{j}]$ is the indicator function taking value 1
if $T_{i}<T_{j}$, otherwise it is 0.

\subsection{The Cox model}

An important survival model used in SurvLIME and SurvNAM is the Cox model
\cite{Cox-1972}. According to this model, the CHF at time $t$ given vector
$\mathbf{x}$ is defined as :
\begin{equation}
H(t|\mathbf{x},\mathbf{b})=H_{0}(t)\exp \left(  \mathbf{b}^{\mathrm{T}%
}\mathbf{x}\right)  , \label{SurvLIME1_10}%
\end{equation}
where $H_{0}(t)$ is a baseline CHF; $\mathbf{b}^{\mathrm{T}}=(b_{1}%
,...,b_{m})$ is a vector of the model parameters in the form of the regression
coefficients which can be found by maximizing the partial likelihood function
for the dataset $\mathcal{A}$ \cite{Cox-1972}.

The Cox model can be rewritten through the SF as
\begin{equation}
S(t|\mathbf{x},\mathbf{b})=\left(  S_{0}(t)\right)  ^{\exp \left(
\mathbf{b}^{\mathrm{T}}\mathbf{x}\right)  }, \label{Cox_SF}%
\end{equation}
where $S_{0}(t)$ is a baseline SF.

\subsection{The Beran estimator}

Another important model which is used in SurvBeX and will be used in SurvBeNIM
is the Beran estimator \cite{Beran-81} which estimates the SF on the basis of
the dataset $\mathcal{A}$ as follows:
\begin{equation}
S_{B}(t|\mathbf{x})=\prod_{t_{i}\leq t}\left \{  1-\frac{\alpha(\mathbf{x}%
,\mathbf{x}_{i})}{1-\sum_{j=1}^{i-1}\alpha(\mathbf{x},\mathbf{x}_{j}%
)}\right \}  ^{\delta_{i}}, \label{Beran_est}%
\end{equation}
where time moments $t_{1},...,t_{n}$ are ordered; the weight $\alpha
(\mathbf{x},\mathbf{x}_{i})$ conforms with relevance of the $i$-th instance
$\mathbf{x}_{i}$ to the vector $\mathbf{x}$ and can be defined by using
kernels as%
\begin{equation}
\alpha(\mathbf{x},\mathbf{x}_{i})=\frac{K(\mathbf{x},\mathbf{x}_{i})}%
{\sum_{j=1}^{n}K(\mathbf{x},\mathbf{x}_{j})}.
\end{equation}

In particular, the use of the Gaussian kernel with the parameter $\tau$ leads
to the following weights $\alpha(\mathbf{x},\mathbf{x}_{i})$:
\begin{equation}
\alpha(\mathbf{x},\mathbf{x}_{i})=\text{\textrm{softmax}}\left(
-\frac{\left \Vert \mathbf{x}-\mathbf{x}_{i}\right \Vert ^{2}}{\tau}\right)  .
\end{equation}

The Beran estimator can also be written in another form:
\begin{equation}
S_{B}(t|\mathbf{x})=\prod_{t_{i}\leq t}\left \{  \frac{1-\sum_{j=1}^{i}%
\alpha(\mathbf{x},\mathbf{x}_{j})}{1-\sum_{j=1}^{i-1}\alpha(\mathbf{x}%
,\mathbf{x}_{j})}\right \}  ^{\delta_{i}}. \label{Beran_est_2}%
\end{equation}

The Beran estimator generalizes the Kaplan-Meier estimator
\cite{Wang-Li-Reddy-2019} which coincides with the Beran estimator when all
weights are identical, i.e., $\alpha(\mathbf{x},\mathbf{x}_{i})=1/n$ for all
$i=1,...,n$.

The CHF by using the Beran estimator is defined as
\begin{align}
H_{B}(t|\mathbf{x})  &  =-\ln S_{B}(t|\mathbf{x})\nonumber \\
&  =\sum_{t_{i}\leq t}\delta_{k}\left[  \ln \left(  1-\sum_{j=1}^{i-1}%
\alpha(\mathbf{x},\mathbf{x}_{j})\right)  -\ln \left(  1-\sum_{j=1}^{i}%
\alpha(\mathbf{x},\mathbf{x}_{j})\right)  \right]  .
\end{align}

\section{General explanation methods}

\subsection{LIME}

LIME is a method for approximating a black-box model $f(\mathbf{x)}$ by a
function $g(\mathbf{x)}$ from a set $G$ of explainable functions, for example,
linear functions, in the vicinity of an explained point $\mathbf{x}$
\cite{Ribeiro-etal-2016}. New instances $\mathbf{z}_{i}$ are generated around
$\mathbf{x}$ and predictions $f(\mathbf{z}_{i}\mathbf{)}$ are computed using
the black-box model. The obtained instances $(\mathbf{z}_{i},f(\mathbf{z}%
_{i}\mathbf{))}$ with weights $v_{i}$ assigned in accordance with distances
between $\mathbf{z}_{i}$ and $\mathbf{x}$ are used to learn the function
$g(\mathbf{x)}$. An explanation (local surrogate) model is trained by solving
the following optimization problem \cite{Ribeiro-etal-2016}:
\begin{equation}
\arg \min_{g\in G}L(f,g,v)+\Phi(g).
\end{equation}

Here $L$ is a loss function, for example, mean squared error (MSE), which
measures how the explanation is close to the prediction of the black-box
model; $\Phi(g)$ is the regularization term.

By having the linear function $g(\mathbf{x)}$, its coefficients explain the
prediction $f(\mathbf{x)}$.

\subsection{A short description of NAM}

\begin{figure}
[ptb]
\begin{center}
\includegraphics[
height=2.2814in,
width=2.4898in
]%
{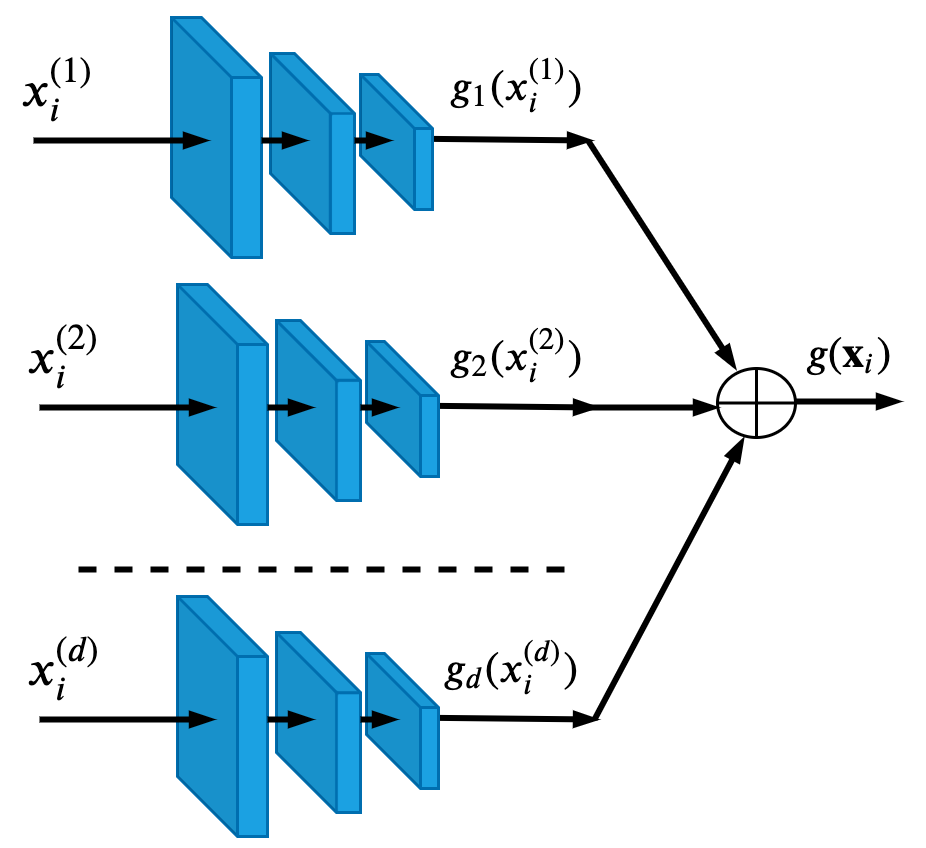}%
\caption{An illustration of NAM}%
\label{fig:NAM_structure}%
\end{center}
\end{figure}

NAM proposed in \cite{Agarwal-etal-21} is implemented as a neural network
consisting of $d$ neural subnetworks (see Fig. \ref{fig:NAM_structure}). Its
architecture is based on applying GAM \cite{Hastie-Tibshirani-1990}, which can
be written as follows:
\begin{equation}
g(\mathbf{x}_{i})=g_{1}(x_{i}^{(1)})+...+g_{d}(x_{i}^{(d)}),
\label{Interpr_GBM_1}%
\end{equation}
where $\mathbf{x}_{i}$ is the feature vector of the $i$-th instance; $g_{i}$
is a univariate shape function; $g$ is a link function.%

A single feature $x_{i}^{(j)}$ of an instance $\mathbf{x}_{i}$ is fed to each
subnetwork. Outcomes of the subnetworks are values of functions $g_{j}%
(x_{i}^{(j)})$, $j=1,...,d$, which can be viewed as impacts of the
corresponding features on the black-box model predictions. If there is a
training set $\mathcal{D}=\{(\mathbf{x}_{1},y_{1}),...,(\mathbf{x}_{n}%
,y_{n})\}$, then the whole network is trained using the following loss
function:
\begin{equation}
L(\mathbf{W},\mathcal{D})=\sum_{i=1}^{n}\left(  y_{i}-\sum_{k=1}^{d}%
g_{k}(x_{i}^{(k)})\right)  ^{2}, \label{SurvNAM_14}%
\end{equation}
where $\mathbf{W}$ is a set of the network training parameters.

It should be noted that we did not consider one of the important methods SHAP
\cite{Lundberg-Lee-2017,Strumbel-Kononenko-2010} because it will not be
directly used in the paper. A detailed description of an extension of SHAP to
survival analysis, called SurvSHAP, can be found in \cite{Krzyzinski-etal-23}.

\section{Explanation methods for survival models}

\subsection{SurvLIME}

SurvLIME is a modification of LIME when the black-box model is a survival
model \cite{Kovalev-Utkin-Kasimov-20a}. Suppose there is a training set
$\mathcal{A}$ and a black-box model whose prediction is represented in the
form of the CHF $H(t|\mathbf{x})$ for every new instance $\mathbf{x}$. In
contrast to the original LIME, SurvLIME approximates the black-box model with
the Cox model, i.e., it approximates the prediction $H(t|\mathbf{x})$ of the
black-box model by the prediction $H_{\text{Cox}}(t|\mathbf{x},\mathbf{b})$ of
the Cox model. Values of parameters $\mathbf{b}$ can be regarded as
quantitative impacts on the prediction $H(t|\mathbf{x})$. They are unknown and
have to be found by means of the approximation of models. Optimal coefficients
$\mathbf{b}$ make the distance between CHFs $H(t|\mathbf{x})$ and
$H_{\text{Cox}}(t|\mathbf{x},\mathbf{b})$ for the instance $\mathbf{x}$ as
small as possible.

In order to implement the approximation at a point $\mathbf{x}$, many nearest
instances $\mathbf{x}_{k}$ are generated in a local area around $\mathbf{x}$.
The CHF $H(t|\mathbf{x}_{k})$ predicted by the black-box model is computed for
every generated $\mathbf{x}_{k}$. Optimal values of $\mathbf{b}$ can be found
by minimizing the weighted average distance between every pair of CHFs
$H(t|\mathbf{x}_{k})$ and $H_{\text{Cox}}(t|\mathbf{x}_{k},\mathbf{b})$ over
all generated points $\mathbf{x}_{k}$. Each weight in the weighted average
distance between CHFs depends on the distance between points $\mathbf{x}_{k}$
and $\mathbf{x}$. Smaller distances between $\mathbf{x}_{k}$ and $\mathbf{x}$
define larger weights of distances between CHFs. The distance metric between
CHFs defines the corresponding optimization problem for computing optimal
coefficients $\mathbf{b}$. Therefore, SurvLIME
\cite{Kovalev-Utkin-Kasimov-20a} uses the $l_{2}$-norm applied to logarithms
of CHFs $H(t|\mathbf{x}_{k})$ and $H_{\text{Cox}}(t|\mathbf{x}_{k}%
,\mathbf{b})$, which leads to a simple convex optimization problem with
variables $\mathbf{b}$.

One of the limitations of SurvLIME is caused by the underlying Cox model: even
in a small area of neighborhoods in the feature space, the proportional
hazards assumption may be violated, leading to an inaccurate surrogate model.
Another limitation is the linear relationship of covariates in the Cox model,
which may be a reason for the inadequate approximation of black-box models by
SurvLIME. In order to overcome the aforementioned SurvLIME drawbacks, it is
proposed to replace the Cox model with the Beran estimator for explaining
predictions produced by a black-box. In other words, it is proposed to
approximate the prediction of the black-box model by means of the Beran
estimator. The corresponding method using this estimator is called SurvBeX
\cite{Utkin-Eremenko-Konstantinov-23}.

\subsection{SurvBeX}

According to SurvBeX \cite{Utkin-Eremenko-Konstantinov-23}, the vector of
parameters $\mathbf{b}=(b_{1},...,b_{d})$ is incorporated into the Beran
estimator (\ref{Beran_est}) as follows:%
\begin{equation}
\alpha(\mathbf{x},\mathbf{x}_{i},\mathbf{b})=\text{\textrm{softmax}}\left(
-\frac{\left \Vert \mathbf{b}\odot \left(  \mathbf{x}-\mathbf{x}_{i}\right)
\right \Vert ^{2}}{\tau}\right)  . \label{Beran_est_32}%
\end{equation}

Here $\mathbf{b}\odot \mathbf{x}$ is the dot product of two vectors. Each
element of $\mathbf{b}$ can be viewed as a quantitative impact of the
corresponding feature on the prediction $S_{B}(t|\mathbf{x},\mathbf{b})$. If
the explained instance is $\mathbf{x}$, then SurvBeX aims to find the vector
$\mathbf{b}$ by approximating the SF $S(t|\mathbf{x})$ of the black-box model
by the SF $S_{B}(t|\mathbf{x},\mathbf{b})$ of the Beran estimator.

For explaining the prediction $S(t|\mathbf{x})$, $N$ points $\mathbf{z}%
_{1},...,\mathbf{z}_{N}$ are generated in a local area around $\mathbf{x}$.
For every point $\mathbf{z}_{k}$, the SF $S(t|\mathbf{z}_{k})$ produced by the
black-box model is computed. Hence, the vector $\mathbf{b}$ can be found by
solving the following optimization problem:%
\begin{equation}
\mathbf{b}^{opt}=\arg \min_{\mathbf{b}}\sum_{j=1}^{N}v_{j}\sum_{i=1}^{n}\left(
S^{(i)}(\mathbf{z}_{j})-S_{B}^{(i)}(\mathbf{z}_{j},\mathbf{b})\right)
^{2}\left(  t_{i}-t_{i-1}\right)  .
\end{equation}
where the weight $v_{j}$ is defined by the distance between vectors
$\mathbf{z}_{j}$ and $\mathbf{x}$; $S_{B}(t|\mathbf{z}_{k},\mathbf{b})$ is
represented by using (\ref{Beran_est}) as a function of the vector
$\mathbf{b}$; $t_{1},...,t_{n}$ are times to the event of interest taken from
the dataset $\mathcal{A}$; $S^{(i)}$ is the value of the SF in the time
interval $[t_{i-1},t_{i})$.

\subsection{SurvNAM}

SurvNAM \cite{Utkin-Satyukov-Konstantinov-22} is an explanation method which
can be viewed as an extension of NAM\  \cite{Agarwal-etal-21} to the case of
survival data. It combines NAM and SurvLIME with the extended Cox model. The
main goal of the NAM-network in SurvNAM is to implement the function
$g(\mathbf{x})=g_{1}(x^{(1)})+...+g_{d}(x^{(d)})$ of features for the Cox
model instead of the product $\mathbf{b}^{\mathrm{T}}\mathbf{x}$.

\begin{figure}
[ptb]
\begin{center}
\includegraphics[
height=3.1349in,
width=5.0868in
]%
{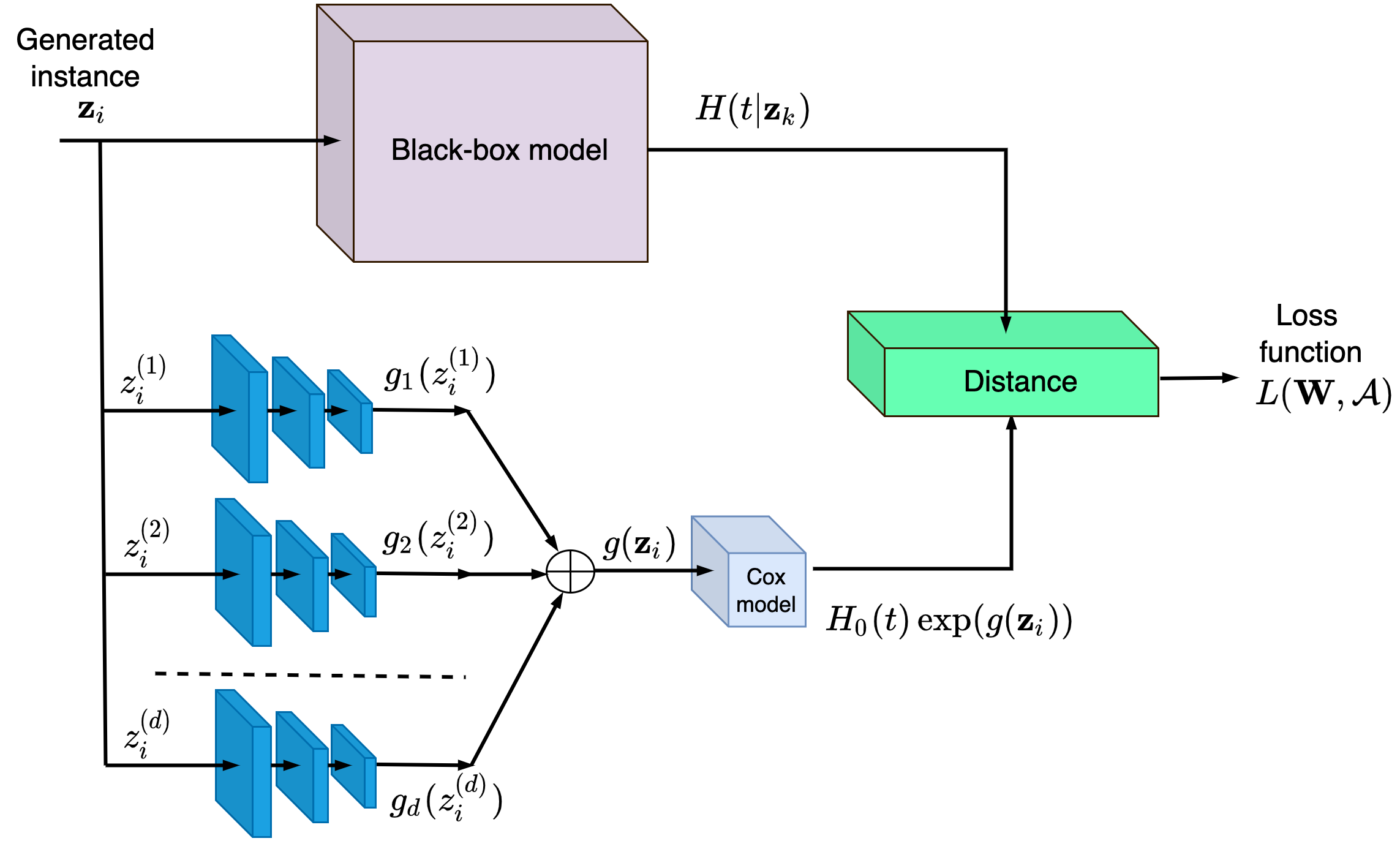}%
\caption{A scheme of the neural network implementing SurvNAM}%
\label{f:survnam}%
\end{center}
\end{figure}

An illustration of SurvNAM is shown in Fig. \ref{f:survnam} where each shape
function $g_{i}(x^{(i)})$ is implemented by means of the $i$-th neural
subnetwork. Then all shape functions are summed to obtain $g(\mathbf{x})$
which is used in the extended Cox model. For explaining the prediction of the
black-box model in the forms of the SF $S(t|\mathbf{x})$ or the CHF
$H(t|\mathbf{x})$, $N$ points $\mathbf{z}_{1},...,\mathbf{z}_{N}$ are
generated in a local area around $\mathbf{x}$. For every point $\mathbf{z}%
_{k}$, the CHF $H(t|\mathbf{z}_{k})$ is produced by the black-box model.
SurvNAM aims to approximate the CHFs $H(t|\mathbf{z}_{k})$ by the CHFs
$H(t|\mathbf{z}_{k},g_{j}(\mathbf{z}_{k},\mathbf{w}_{j}%
),j=1,...,d)=H(t|\mathbf{z}_{k},g(\mathbf{z}_{k},\mathbf{W})$, $k=1,...,N$.
Here $\mathbf{w}_{j}\subset \mathbf{W}$ is the subset of weights of the $k$-th
subnetwork, $\mathbf{w}_{1}\cup...\cup \mathbf{w}_{m}=\mathbf{W}$. This implies
that the loss function for training the neural network (NAM) should be
represented as a distance between $H(t|\mathbf{z}_{k})$ and $H(t|\mathbf{z}%
_{k},g(\mathbf{z}_{k},\mathbf{W}))$ or between their logarithms:%
\begin{equation}
L(\mathbf{W},\mathcal{A})=\sum_{i=1}^{N}\sum_{j=0}^{n}v_{i}\left(
H_{j}(\mathbf{z}_{i})-H_{0j}\exp \left(  g(\mathbf{z}_{i},\mathbf{W})\right)
\right)  ^{2}\left(  t_{j}-t_{j-1}\right)  , \label{SurvNAM_1}%
\end{equation}
where $H_{j}$ and $H_{0j}$ are values of $H(t|\mathbf{z})$ and of the baseline
CHF $H_{0}(t)$, respectively, in the time interval $[t_{i-1},t_{i})$;
$t_{1},...,t_{n}$ are times to the event of interest taken from the dataset
$\mathcal{A}$; $H_{0j}\exp \left(  g(\mathbf{z}_{i},\mathbf{W}\right)  $ is the
expression for the Cox model in the time interval $[t_{i-1},t_{i})$; $v_{i}$
is the weight of the $i$-th generated instance defined by the distance between
vectors $\mathbf{z}_{i}$ and $\mathbf{x}$.

The loss function can be also represented through logarithms of CHFs as
follows:
\begin{equation}
L(\mathbf{W},\mathcal{A})=\sum_{i=1}^{N}\sum_{j=0}^{n}v_{i}\left(  \ln
H_{j}(\mathbf{z}_{i})-\ln H_{0j}-g(\mathbf{z}_{i},\mathbf{W})\right)
^{2}\left(  t_{j}-t_{j-1}\right)  . \label{SurvNAM_2}%
\end{equation}

It should be noted that the loss functions (\ref{SurvNAM_1}) and
(\ref{SurvNAM_2}) are different. However, it is shown in
\cite{Utkin-Satyukov-Konstantinov-22} that the function (\ref{SurvNAM_2}) can
also be used with some conditions because it simplifies the neural network
training process.%

\section{The proposed SurvBeNIM method}

SurvBeNIM is a combination of SurvBeX \cite{Utkin-Eremenko-Konstantinov-23}
and NAM \cite{Agarwal-etal-21} or SurvNAM
\cite{Utkin-Satyukov-Konstantinov-22}. However, in contrast to NAM or SurvNAM,
SurvBeNIM does not consider the shape function. SurvBeNIM aims to find an
\emph{importance function} $h_{j}(x^{(j)})$ of the $j$-th feature, which can
be viewed as a function whose value at each $x^{(j)}$ shows how $x^{(j)}$
impacts on the ground truth function $f(\mathbf{x})=f(x^{(1)},\dots,x^{(d)})$,
i.e. $h_{j}(x^{(j)})$ can be regarded as a quantitative measure of the feature
$x^{(j)}$ impact on $f(\mathbf{x})$.

We propose to incorporate the importance functions $h_{j}(x^{(j)})$,
$j=1,...,d$, into the kernels of the Beran estimator. The normalized kernel
$\alpha(\mathbf{x},\mathbf{x}_{i},\mathbf{g})$ can be rewritten as follows:
\begin{equation}
\alpha(\mathbf{x},\mathbf{x}_{i},\mathbf{g})=\text{\textrm{softmax}}\left(
-\tau^{-1}\sum_{j=1}^{d}h_{j}(x^{(j)})\left(  x^{(j)}-x_{i}^{(j)}\right)
^{2}\right)  . \label{attent_weight_Beran}%
\end{equation}

It should be noted that the function $h_{j}(x^{(j)})$ cannot be viewed as the
shape function because, according to GAM, the shape function reflects the
dependence of $f$ on the feature $x^{(j)}$ in a target domain $\mathcal{Y}$
and its change determines the importance of the corresponding feature. In
contrast to the shape function, $h_{j}(x^{(j)})$ in (\ref{attent_weight_Beran}%
) is an average weight of the $j$-th feature or its importance. To be more
rigorous, the importance function weighs the distance between the $j$-th
feature of the explained instance $\mathbf{x}$ and training instances
$\mathbf{x}_{i}$. Therefore, the importance function in SurvBeNIM directly
depends on the training set whereas the shape function in SurvNAM implicitly
depends on the training set through the baseline function which may be
incorrect when the analyzed dataset has a complex structure.

Let us consider the extended Beran estimator which is used with the importance
functions implemented by means of $d$ neural subnetworks in the same way as it
is realized in SurvNAM. For the local explanation, a set $\mathcal{Z}$ of $N$
instances $\mathbf{z}_{1},...,\mathbf{z}_{N}$ are randomly generated around
the vector $\mathbf{x}$ in accordance with some probability distribution, for
example, with the multivariate normal distribution. The generated instances
are fed to the black-box survival model to obtain SFs $S(t|\mathbf{z}%
_{1}),...,S(t|\mathbf{z}_{N})$ or CHFs $H(t|\mathbf{z}_{1}),...,H(t|\mathbf{z}%
_{N})$. Similarly to SurvNAM, we use the weighted mean distance between SFs or
CHFs of the black-box survival model and the neural implementation of the
extended Beran estimator as a loss function. Let $S^{(i)}(\mathbf{x}%
,\mathbf{h}(\mathbf{z},\mathbf{W}))$ be a value of $S(t|\mathbf{x}%
,\mathbf{\ h}(\mathbf{z},\mathbf{W}))$ in the time interval $[t_{i-1},t_{i})$
under condition of using the function \textbf{$h$}$(\mathbf{x},\mathbf{W})$
implemented by $d$ neural subnetworks and defined as%
\begin{equation}
\mathbf{h}(\mathbf{x},\mathbf{W})=\left(  h_{1}(x^{(1)},\mathbf{w}%
_{1}),...,h_{d}(x^{(d)},\mathbf{w}_{d})\right)  ,
\end{equation}
where $\mathbf{w}_{1},...,\mathbf{w}_{d}$ are sets of subnetwork weights,
$\mathbf{w}_{1}\cup...\cup \mathbf{w}_{m}=\mathbf{W}$.

Using the expression for the Beran estimator (\ref{Beran_est_2}), we can write
the approximating SF in the time interval $[t_{i-1},t_{i})$ as follows:
\begin{align}
&  S_{B}^{(i)}(\mathbf{x}_{j},\mathbf{h}(\mathbf{z}_{j},\mathbf{W}%
))\nonumber \\
&  =\prod_{t_{s}\leq t_{i}}\left \{  \frac{1-\sum_{j=1}^{s}%
\text{\textrm{softmax}}\left(  -A(\mathbf{x}_{i},\mathbf{h}(\mathbf{z}%
_{j},\mathbf{W)}\right)  }{1-\sum_{j=1}^{s-1}\text{\textrm{softmax}}\left(
-A(\mathbf{x}_{i},\mathbf{h}(\mathbf{z}_{j},\mathbf{W)}\right)  }\right \}
^{\delta_{s}},
\end{align}
where
\begin{equation}
A(\mathbf{x}_{i},\mathbf{h}(\mathbf{z}_{j},\mathbf{W})\mid \tau)=\tau^{-1}%
\sum_{k=1}^{d}h_{k}(z_{j}^{(k)},\mathbf{w}_{k})\left(  x_{i}^{(k)}-z_{j}%
^{(k)}\right)  ^{2}. \label{A_meth1}%
\end{equation}

We can consider two strategies of the neural network training. The first one
is to train the network only for the explained instance. It implements the
local explanation. According to this strategy, $N$ instances are randomly
generated around the vector $\mathbf{x}$, and the network is trained on the
basis of these instances. In this case, the network cannot explain an instance
different from $\mathbf{x}$. Nevertheless, the network explains the given
explained instance and allows us to study how its features impact on the
predicted SF or CHF. The loss function for the first strategy is of the form:
\begin{equation}
L(\mathbf{W},\mathbf{x},\mathcal{Z})=\sum_{j=1}^{N}v_{j}\sum_{i=1}^{n}\left(
S^{(i)}(\mathbf{z}_{j})-S_{B}^{(i)}(\mathbf{x},\mathbf{h}(\mathbf{z}%
_{j},\mathbf{W}))\right)  ^{2}\left(  t_{i}-t_{i-1}\right)  ,
\label{Loss_SurvBeNAM_1}%
\end{equation}
where $v_{j}$ is a weight which depends on the distance between the explained
point $\mathbf{x}$ and generated points $\mathbf{z}_{j}$.

It is important to note that, in contrast to the function $g_{j}(x^{(j)})$ in
SurvNAM, $h_{j}(x^{(j)})$ does not represent the shape function, it represents
directly the feature importance. The importance function contains the total
information to explain the corresponding prediction.

\begin{figure}
[ptb]
\begin{center}
\includegraphics[
height=3.2621in,
width=5.9897in
]%
{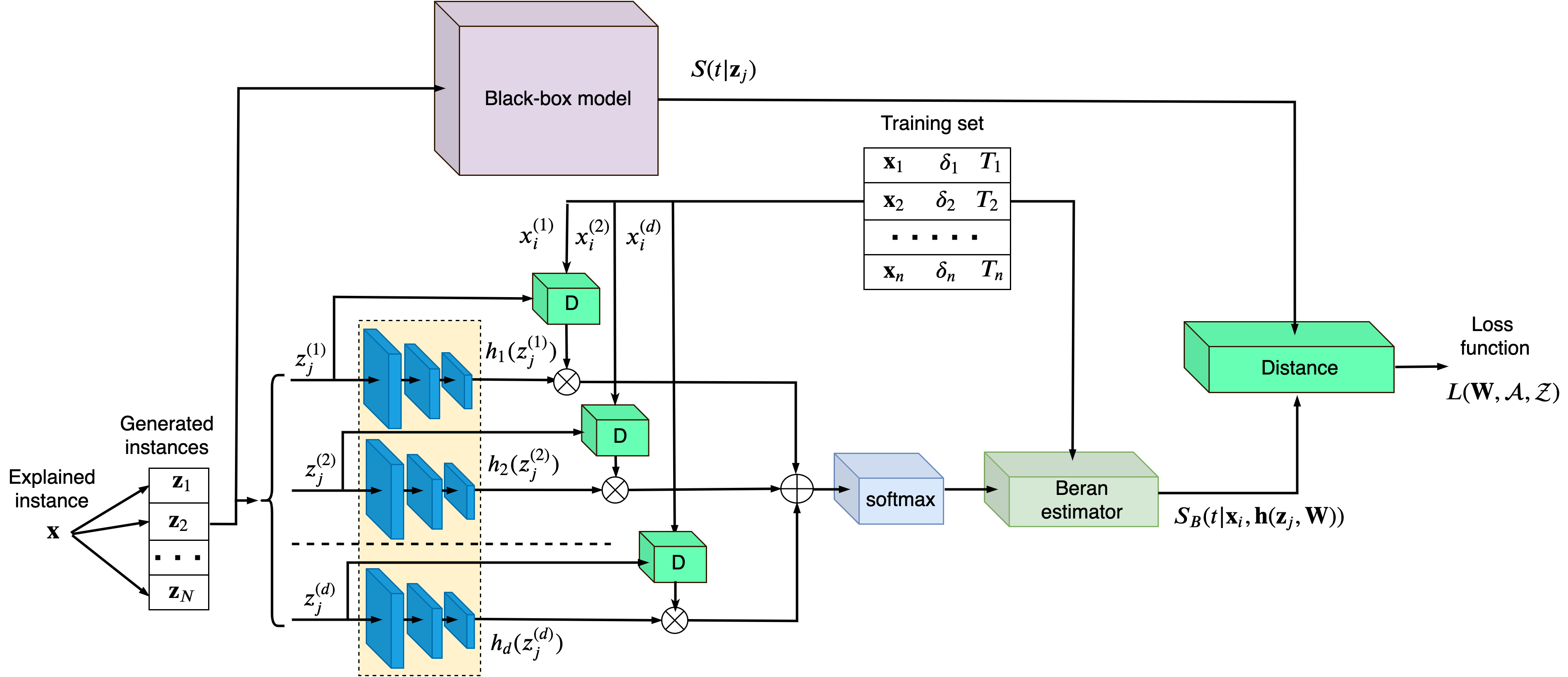}%
\caption{A general scheme illustrating SurvBeNIM}%
\label{f:survbenam}%
\end{center}
\end{figure}

A general scheme illustrating SurvBeNIM is depicted in Fig.\ref{f:survbenam}
where the block \textquotedblleft D\textquotedblright \ computes the distance
between features $x^{(j)}-z_{i}^{(j)}$. Outputs of the softmax operations are
weights $\alpha(\mathbf{x},\mathbf{z}_{j},\mathbf{h})$, $j=1,...,d$, which are
fed to the block \textquotedblleft Beran estimator\textquotedblright%
\ producing the prediction in the form of the SF $S_{B}(t|\mathbf{x}%
,\mathbf{h}(\mathbf{z}_{j},\mathbf{W)})$. The loss function computes the
distance between SFs $S_{B}(t|\mathbf{x},\mathbf{h}(\mathbf{z}_{j}%
,\mathbf{W)})$ and $S(t|\mathbf{z}_{j})$. The scheme illustrates the training
stage of SurvNAM. The inference stage uses only subnetworks in the rectangle
with the dashed bounds, which compute functions $h_{i}(z^{(i)},\mathbf{w}%
_{i})$, $i=1,...,d$.%

The second strategy tries to train a neural network which can be used for
explaining instances without retraining. In order to implement this strategy,
we generate $N$ instances around each training instance $\mathbf{x}_{r}$,
$r=1,...,n$. As a result, we obtain $N\cdot n$ generated instances, denoted as
$\mathbf{z}_{j,r}$, $j=1,...,N$, $r=1,...,n$. We use the additional index $r$
for generated points $\mathbf{z}_{j,r}$, which indicates that the points were
generated around the point $\mathbf{x}_{r}\in \mathcal{A}$. Let $v_{j,r}$ be a
weight depending on the distance between the point $\mathbf{x}_{r}$ and the
generated point $\mathbf{z}_{j,r}$. Finally, the loss function for the
explained instance $\mathbf{x}_{r}$ is of the form:%
\begin{equation}
L_{r}(\mathbf{W},\mathbf{x}_{r},\mathcal{Z})=\sum_{j=1}^{N}v_{j,r}\sum
_{i=1}^{n}\left(  S^{(i)}(\mathbf{z}_{j,r})-S_{B}^{(i)}(\mathbf{x}%
_{r},\mathbf{h}(\mathbf{z}_{j,r},\mathbf{W}))\right)  ^{2}\left(
t_{i}-t_{i-1}\right)  . \label{Loss_SurvBeNAM}%
\end{equation}

By using the entire dataset for training the neural network, we can write the
final loss function taking into account all points from $\mathcal{A}$:
\begin{equation}
L(\mathbf{W},\mathcal{A},\mathcal{Z})=\sum_{r=1}^{n}L_{r}(\mathbf{W}%
,\mathbf{x}_{r}). \label{Loss_SurvBeNAM_full}%
\end{equation}

It can be seen from the above, that the second strategy requires more time for
training the neural network, but it allows us to avoid the repeated training
for different instances. In fact, this strategy implements the global explanation.

One of the difficulties of using the distance between SFs is possible small or
large values of SFs. In order to overcome this difficulty, it is proposed to
introduce additional weights for values of SFs. Let us find the expectation of
the random time to the event of interest by means of the SF predicted by the
black-box model. It can be obtained as
\begin{equation}
m_{j}=\sum_{i=1}^{n}\left(  t_{i}-t_{i-1}\right)  \cdot S^{(i)}(\mathbf{z}%
_{j}).
\end{equation}

We assume that values of $S(t\mid \mathbf{z}_{j})$ for values of $t$, which are
close to $m_{j}$, are more important in comparison with values of the SF for
values $t$, which are far from $m_{j}$. Relying on this assumption,
(\ref{Loss_SurvBeNAM_1}) can be rewritten as follows:
\begin{equation}
L(\mathbf{W},\mathbf{x},\mathcal{Z})=\sum_{j=1}^{N}v_{j}\sum_{i=1}^{n}%
p_{i,j}\cdot \left(  S^{(i)}(\mathbf{z}_{j})-S_{B}^{(i)}(\mathbf{x}%
,\mathbf{h}(\mathbf{z}_{j},\mathbf{W}))\right)  ^{2}\left(  t_{i}%
-t_{i-1}\right)  ,
\end{equation}

Here $p_{i,j}$ is defined as
\begin{equation}
p_{i,j}=K^{\ast}\left(  m_{j},\frac{t_{i}+t_{i-1}}{2}\right)  ,
\end{equation}
where $K^{\ast}$ is the normalized kernel which measures how the expected
value $m_{j}$ is close to the center of the interval $[t_{i-1},t_{i})$.

In the case of the Gaussian kernel, we can write%
\begin{equation}
h_{i,j}=\text{\textrm{softmax}}\left(  -\frac{\left(  m_{j}-(t_{i}%
+t_{i-1})/2\right)  ^{2}}{\varkappa}\right)  ,
\end{equation}
where $\varkappa$ is the kernel parameter.

The above trick is heuristic, but it can improve the explanation procedure.

\section{Numerical experiments}

\subsection{Numerical experiments with synthetic data}

\subsubsection{Data generation parameters}

Similarly to \cite{Utkin-Eremenko-Konstantinov-23}, we use synthetic data in
the form of $n$ triplets $(\mathbf{x}_{i},\delta_{i},T_{i}),i=1,...,n$, where
random survival times $T_{i}$ are generated in accordance with the Cox model
having the predefined vector $\mathbf{b}^{true}$ of parameters to compare the
obtained measures provided by different models of the feature impact with a
groundtruth vector $\mathbf{b}^{true}$. An approach to generate the Cox model
survival times proposed in \cite{Bender-etal-2005} is used. According to this
approach, the survival times are governed by the Weibull distribution with the
scale $\lambda=10^{-5}$ and shape $v=2$ parameters and determined as follows
\cite{Bender-etal-2005}:
\begin{equation}
T=\left(  \frac{-\ln(U)}{\lambda \exp(\mathbf{b}^{true}\mathbf{x}^{\mathrm{T}%
})}\right)  ^{1/v}, \label{Cox_generator}%
\end{equation}
where $U$ is a random value uniformly distributed from interval $[0,1]$;
$\mathbf{b}^{true}=(b_{1}^{true},...,b_{d}^{true})$ is the predefined vector
of parameters that are set in advance.

In order to make the data structure more complex, we generate two clusters of
feature vectors with centers $\mathbf{p}_{i}=(p_{i1},...,p_{id})$, $i=1,2$,
the radius $R$, and different vectors of the Cox model coefficients
$\mathbf{b}_{1}^{true}$ and $\mathbf{b}_{2}^{true}$. Parameters of the
clusters are $d=5$, $\mathbf{b}_{1}^{true}=(0.5,0.25,0.12,0,0)$,
$\mathbf{b}_{2}^{true}=(0,0,0.12,0.25,0.5)$, $\mathbf{p}_{1}%
=(0.25,0.25,0.25,0.25,0.25)$, $\mathbf{p}_{2}=(0.75,0.75,0.75,0.75)$, $R=0.2$.
The number of points in each cluster is $200$. In the same way, the point
consisting of $d=20$ features are generated such that $\mathbf{b}_{1}%
^{true}=(0.5,0.25,0.12,0,0,...,0)$, $\mathbf{b}_{2}^{true}%
=(0,...,0,0.12,0.25,0.5)$.

We also generate five clusters of feature vectors of size $d=10$ with
parameters
\begin{align}
\mathbf{b}_{1}^{true}  &  =(0,0,0.4,0.8,0,0,0,0,0,0),\nonumber \\
\mathbf{b}_{2}^{true}  &  =(0,0,0.8,0,0,0,0,0.4,0,0),\nonumber \\
\mathbf{b}_{3}^{true}  &  =(0,0,0,0.4,0,0.8,0,0,0,0),\nonumber \\
\mathbf{b}_{4}^{true}  &  =(0,0.4,0.8,0,0,0,0,0,0,0),\nonumber \\
\mathbf{b}_{5}^{true}  &  =(0,0,0,0,0,0,0.8,0,0.4,0),
\end{align}%
\begin{equation}
p_{ij}=0.1+0.2\cdot(i-1),\ i=1,...,5,\ j=1,...,10.
\end{equation}

We use the RSF \cite{Ibrahim-etal-2008} consisting of $100$ decision survival
trees as a black-box model with $\sqrt{d}$ features selected to build each
tree in the RSF and the largest depth of each tree equal to $8$. The log-rank
splitting rule is used.

The number of generated points around an explained instance $\mathbf{x}$ is
$N=100$. The points are generated in accordance with the normal distribution
such that its expectation is the explained point $\mathbf{x}$ and the standard
deviation is $0.2$. Weights of generated instances $w_{j}$ are assigned by
using the Gaussian kernel with the width parameter $\sigma=0.4$.

In experiments with the global explanation, we use the training set for
learning the RSF as a black box, then the testing set with perturbed points
around every testing point are used for learning the SurvNAM and SurvBeNIM.

\subsubsection{Measures for comparison}

Different explanation methods are compared by means of the following
measures:
\begin{equation}
D(\mathbf{b}^{{model}},\mathbf{b}^{true})=\left \Vert \mathbf{b}^{{model}%
}-\mathbf{b}^{true}\right \Vert ^{2},
\end{equation}%
\begin{equation}
KL(\mathbf{b}^{{model}},\mathbf{b}^{true})=\sum_{i=1}^{n}b_{i}^{true}\ln
\frac{b_{i}^{true}}{b_{i}^{{model}}},
\end{equation}%
\begin{equation}
C(\mathbf{b}^{{model}},\mathbf{b}^{true})=\frac{\sum \nolimits_{i,j}%
\mathbf{1}[b_{i}^{true}<b_{j}^{true}]\cdot \mathbf{1}[b_{i}^{{model}}%
<b_{j}^{{model}}]}{\sum \nolimits_{i,j}\mathbf{1}[b_{i}^{true}<b_{j}^{true}]}.
\end{equation}

It can be seen from the above that $D(\mathbf{b}^{{model}},\mathbf{b}^{true})$
is the Euclidean distance between the two vectors $\mathbf{b}^{{model}}$ and
$\mathbf{b}^{true}$ which are normalized. The vector $\mathbf{b}^{{model}}$ is
obtained by the corresponding explanation model. The normalized vectors
$\mathbf{b}^{{model}}$ and $\mathbf{b}^{true}$ can be viewed as discrete
probability distributions, therefore, the Kullback--Leibler divergence
$KL(\mathbf{b}^{{model}},\mathbf{b}^{true})$ can also be used to compare the
methods. $C(\mathbf{b}^{{model}},\mathbf{b}^{true})$ is the C-index which
estimates the probability that elements of vector $\mathbf{b}^{{model}}$ are
ranking in the same way as elements of vector $\mathbf{b}^{true}$. The C-index
is the most important because the vectors $\mathbf{b}^{{model}}$ and
$\mathbf{b}^{true}$ may be different, but their elements are identically
ranked in the \textquotedblleft ideal\textquotedblright \ case of the
explanation. The explanation aims to select the most important features, i.e.
it aims to analyze the relationship between features. The C-index measures
this relationship for two vectors. At the same time, it is interesting to
consider also the measures $D(\mathbf{b}^{{model}},\mathbf{b}^{true})$ and
$KL(\mathbf{b}^{{model}},\mathbf{b}^{true})$.

In contrast to SurvLIME and SurvBeX where each feature is explained by the
point-valued measure $b_{i}$, $i=1,...,d$, SurvNAM and SurvBeNIM provide
functions $h$ and $g$, respectively. The vector $\mathbf{b}^{{model}}$ in
SurvBeNIM is determined by averaging of values $h_{1}(x^{(1)}),...,h_{d}%
(x^{(d)})$ over all $\mathbf{z}_{1},...,\mathbf{z}_{N}$ randomly generated
around the vector $\mathbf{x}$. The vector $\mathbf{b}^{{model}}$ in SurvNAM
is computed as the standard deviation of the shape function $g(\mathbf{x})$ in
the vicinity of point $\mathbf{x}$.

It should be noted that the loss functions in the compared explanation methods
are based on minimizing the distance between the surrogate model SF and the
black-box model SF. Therefore, we also consider the following distance
measure:
\begin{equation}
D(S,S_{B-B})=\sum_{i=1}^{n}\left(  S^{(i)}(\mathbf{x})-S_{B-B}^{(i)}%
(\mathbf{x})\right)  ^{2}\left(  t_{i}-t_{i-1}\right)  ,
\end{equation}
where $S^{(i)}(\mathbf{z})$ is the SF predicted by the explanation model for
the explained point $\mathbf{x}$ in the time interval $[t_{i-1},t_{i}]$;
$S_{B-B}^{(i)}$ is the SF predicted by the black-box model in the same time interval.

In order to obtain the mean measures $D(\mathbf{b}^{{model}},\mathbf{b}%
^{true})$, $KL(\mathbf{b}^{{model}},\mathbf{b}^{true})$, $C(\mathbf{b}%
^{{model}},\mathbf{b}^{true})$, and $D(S,S_{B-B})$ computed over a set of $M$
tested points, we use the mean squared distance (MSD) as an analog of the mean
squared error, the mean KL-divergence (MKL), the mean C-index (MCI), and the
mean SF distance (MSFD). For example, the MSD is determined as
\begin{equation}
MSD=\frac{1}{M}\sum_{i=1}^{M}D(\mathbf{b}_{i}^{{model}},\mathbf{b}^{true}),
\end{equation}
where $\mathbf{b}_{i}^{{model}}$ is the vector obtained for the $i$-th tested point.

The greater the value of the MCI and the smaller the MSFD, the MKL, and the
MSD, the better results we get.

All these measures have been used in analyzing SurvBeX
\cite{Utkin-Eremenko-Konstantinov-23}.

\subsubsection{Local explanation: Experiments with synthetic data}

It is interesting to study SurvBeNIM by considering the complex datasets
consisting of several clusters. First, we study the case of two clusters. The
clusters are generated in accordance with the Cox model, but they have quite
different parameters $\mathbf{b}_{1}^{true}$, $\mathbf{b}_{2}^{true}$,
$\mathbf{p}_{1}$, and $\mathbf{p}_{2}$ of the model. This data structure
violates the homogeneous Cox model data structure.%

\begin{figure}
[ptb]
\begin{center}
\includegraphics[
height=5in,
width=5in
]%
{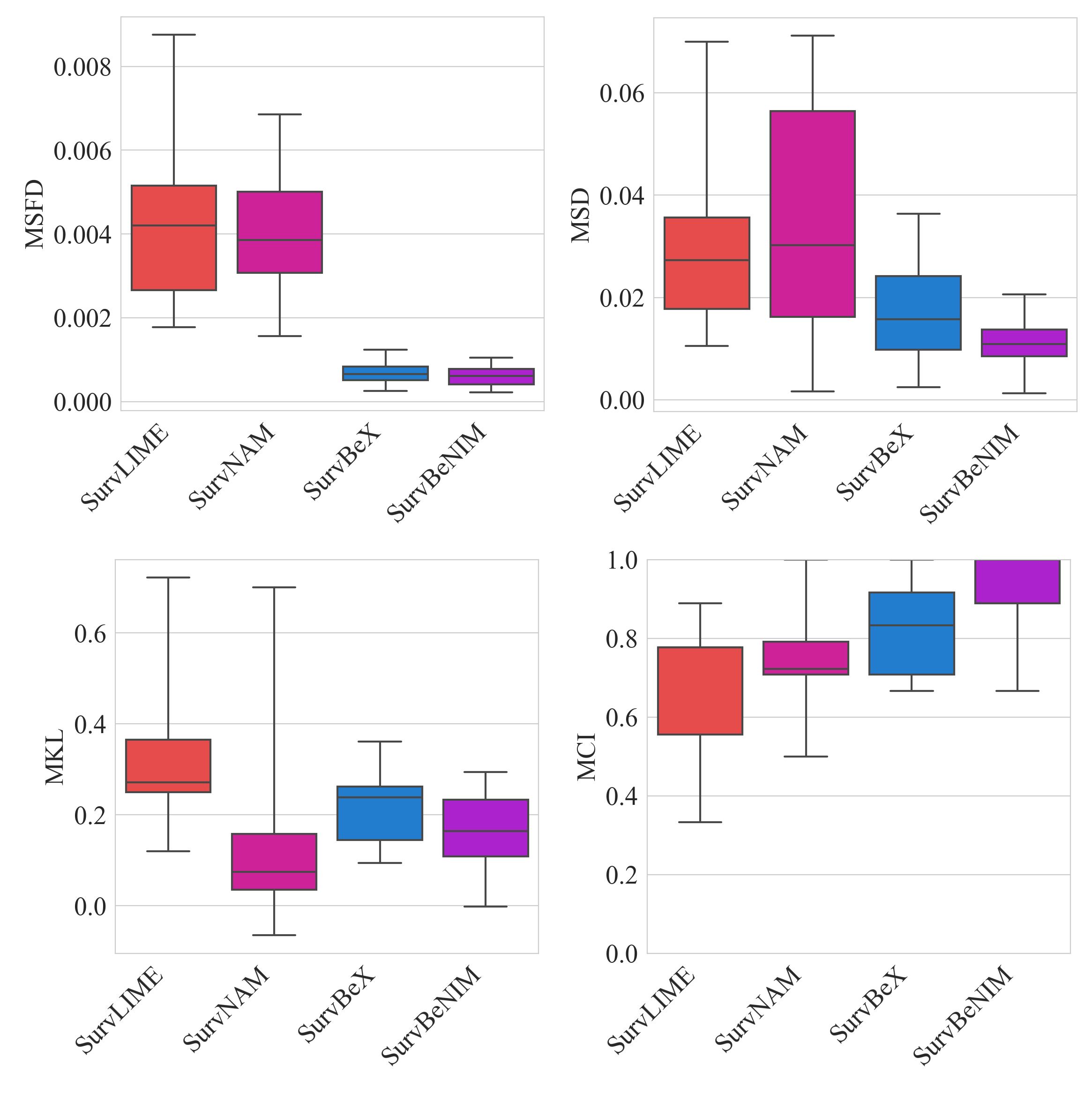}%
\caption{Boxplots illustrating difference between MSFD, MSD, MKL, and MCI for
SurvLIME, SurvNAM, SurvBeX, and SurvBeNIM when the black box (the RSF) is
trained on two clusters of data with 5 features}%
\label{f:bb_c2_f5}%
\end{center}
\end{figure}

Fig. \ref{f:bb_c2_f5} shows boxplots of MSFD, MSD, MKL, and MCI for the
explanation models SurvLIME, SurvNAM, SurvBeX, and SurvBeNIM when the black
box is the RSF and the number of features is 5. It can be seen from Fig.
\ref{f:bb_c2_f5} that SurvBeNIM provides better results in comparison with
other models for all measures except for the MKL measure which shows that
SurvNAM provides a better result. The most important measure MCI shows that
SurvBeNIM clearly outperforms all other explanation models.

\begin{figure}
[ptb]
\begin{center}
\includegraphics[
height=2.4543in,
width=5.3134in
]%
{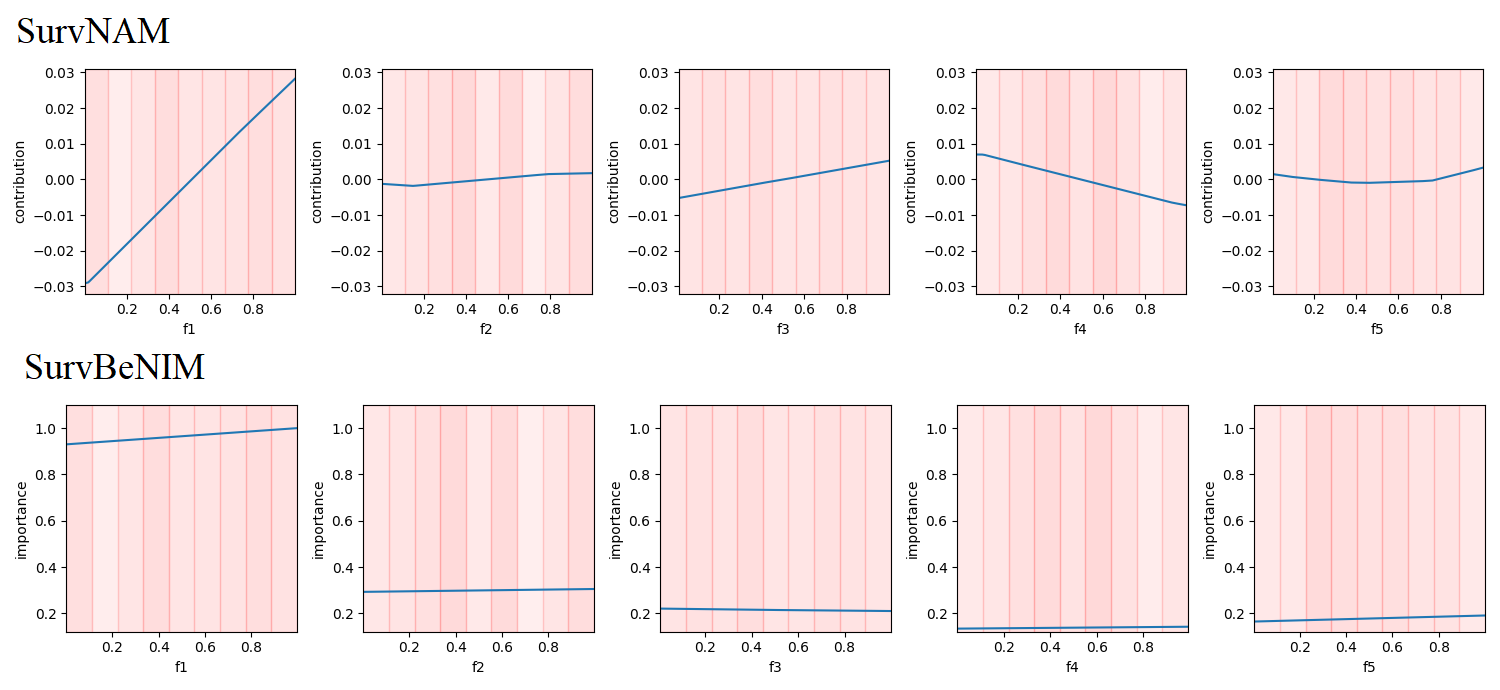}%
\caption{Shape functions obtained by SurvNAM and importance functions obtained
by SurvBeNIM by randomly taking a point from the first cluster when the Cox
model with $5$ features is used for generating the dataset}%
\label{f:c2f5_shape_imp_1}%
\end{center}
\end{figure}

\begin{figure}
[ptb]
\begin{center}
\includegraphics[
height=2.4111in,
width=5.3134in
]%
{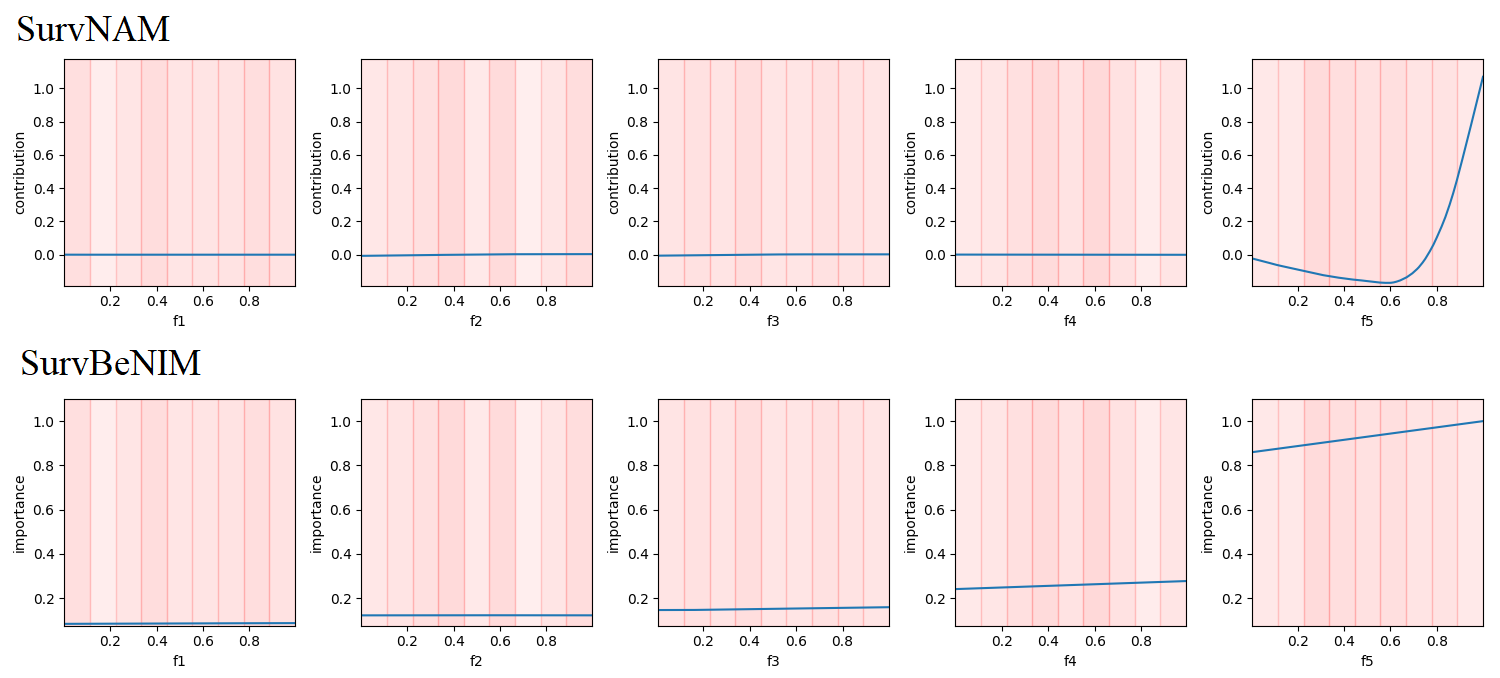}%
\caption{Shape functions obtained by SurvNAM and importance functions obtained
by SurvBeNIM by randomly taking a point from the second cluster when the Cox
model with $5$ features is used for generating the dataset}%
\label{f:c2f5_shape_imp_68}%
\end{center}
\end{figure}

Shape functions for all features obtained by means of SurvNAM and importance
functions obtained by means of SurvBeNIM for two randomly selected points from
two clusters are shown in Figs. \ref{f:c2f5_shape_imp_1} and
\ref{f:c2f5_shape_imp_68}. Since the clusters are respectively defined by two
vectors $\mathbf{b}_{1}^{true}=(0.5,0.25,0.12,0,0)$, $\mathbf{b}_{2}%
^{true}=(0,0,0.12,0.25,0.5)$ of the Cox model, then the important features are
different for these points. It can be seen from Fig. \ref{f:c2f5_shape_imp_1}
that results provided by SurvNAM and SurvBeNIM are similar, i.e., the rapidly
changing shape functions obtained by using SurvNAM correspond to the larger
values of the importance functions provided by SurvBeNIM. At the same time,
results for the second point depicted in Fig. \ref{f:c2f5_shape_imp_68} are
different. One can see from Fig. \ref{f:c2f5_shape_imp_68} that the third and
fourth features are not important in accordance with SurvNAM whereas they are
important in accordance with SurvBeNIM. This implies that SurvBeNIM provides
more correct results.%

%

\begin{figure}
[ptb]
\begin{center}
\includegraphics[
height=5in,
width=5in
]%
{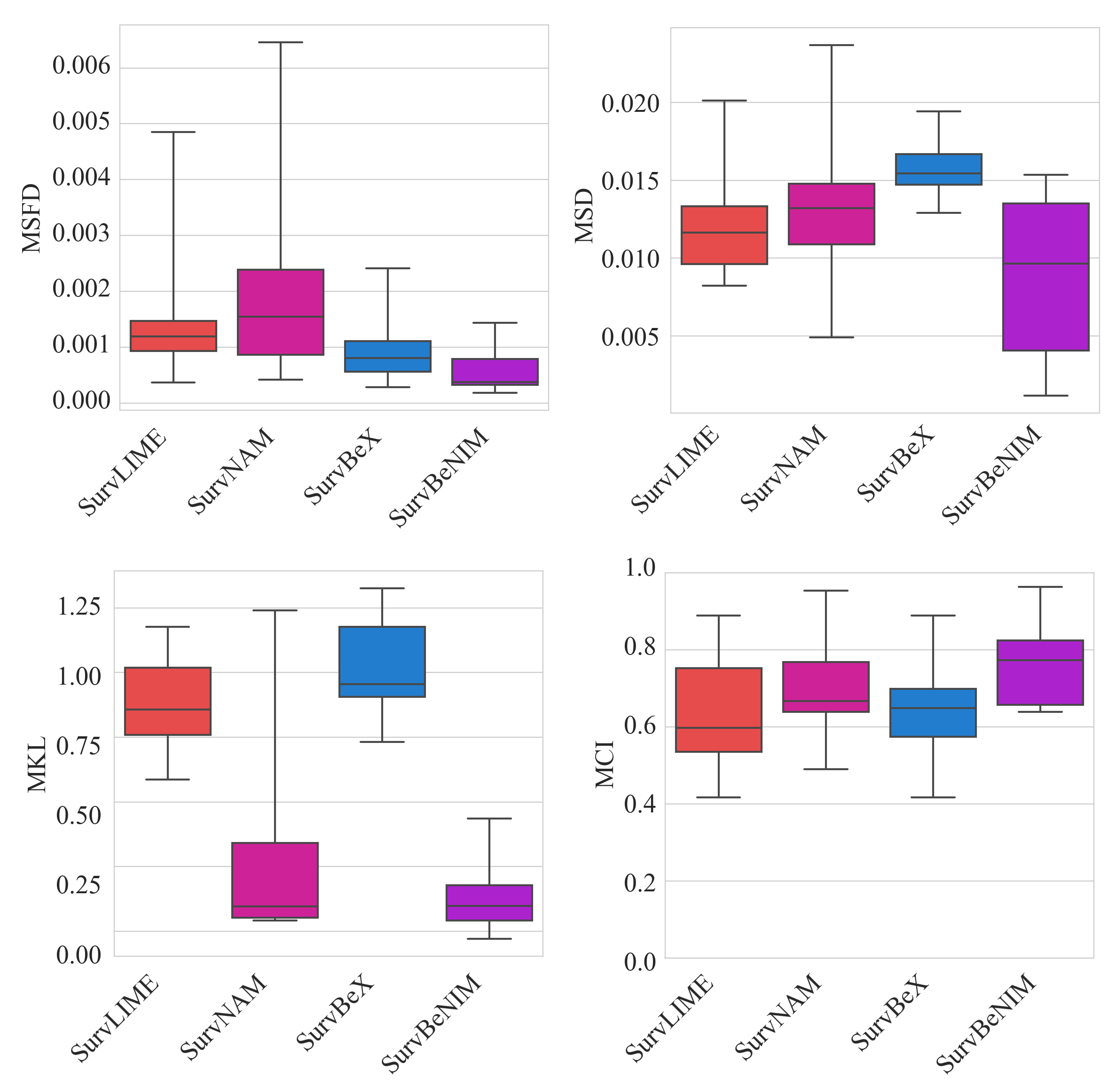}%
\caption{Boxplots illustrating difference between MSFD, MSD, MKL, and MCI for
SurvLIME, SurvNAM, SurvBeX, and SurvBeNIM when the black box (the RSF) is
trained on two clusters of data with 20 features}%
\label{f:bb_c2_f20}%
\end{center}
\end{figure}

Boxplots of the same measures, when the number of features is 20, are depicted
in Fig. \ref{f:bb_c2_f20}. It can also be seen from Fig. \ref{f:bb_c2_f20}
that SurvBeNIM outperforms all other explanation models. Moreover, one can see
from Figs. \ref{f:bb_c2_f5} and \ref{f:bb_c2_f20} that SurvNAM is unstable. It
has a large variance of results in comparison with other models.%

\begin{figure}
[ptb]
\begin{center}
\includegraphics[
height=5in,
width=5in
]%
{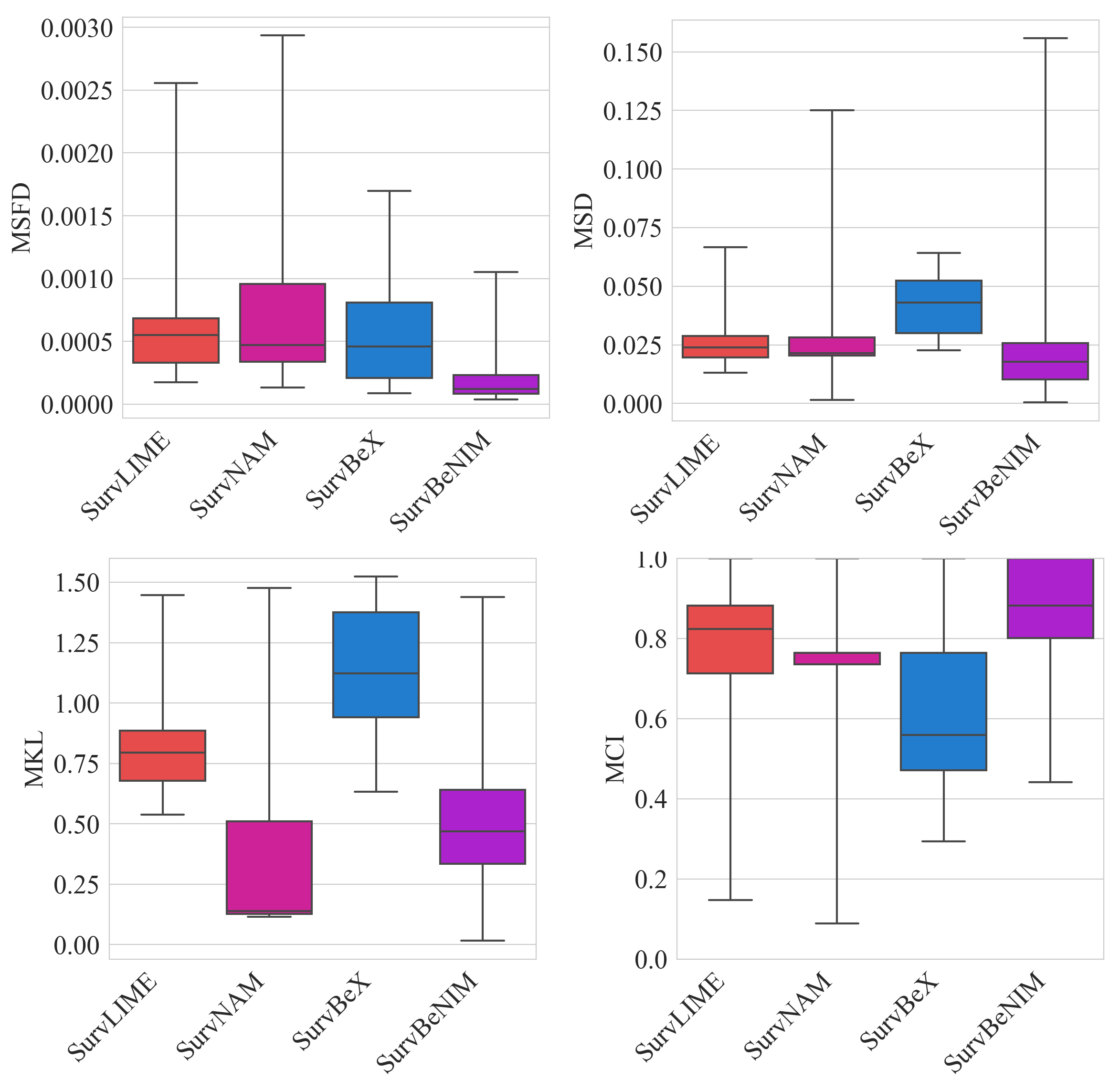}%
\caption{Boxplots illustrating difference between MSFD, MSD, MKL, and MCI for
SurvLIME, SurvNAM, SurvBeX, and SurvBeNIM when the black box (the RSF) is
trained on five clusters of data with 10 features}%
\label{f:bb_c5_f10}%
\end{center}
\end{figure}

We also provide boxplots of the same measures for the case of 5 clusters and
10 features. The corresponding results are depicted in Fig. \ref{f:bb_c5_f10}.
It is interesting to note that SurvBeX provides worse results in the case of 5
clusters. It can be explained by a complex data structure which is formed by
the large number of clusters. This structure cannot be accurately modeled by
means of the Gaussian kernels which are used in SurvBeX. At the same time, the
neural implementation of the kernels allows us to obtain adequate kernels with
many training parameters (weights of the network connections) which are
sensitive to the complex structure of data.%

\begin{figure}
[ptb]
\begin{center}
\includegraphics[
height=4.1537in,
width=4.7279in
]%
{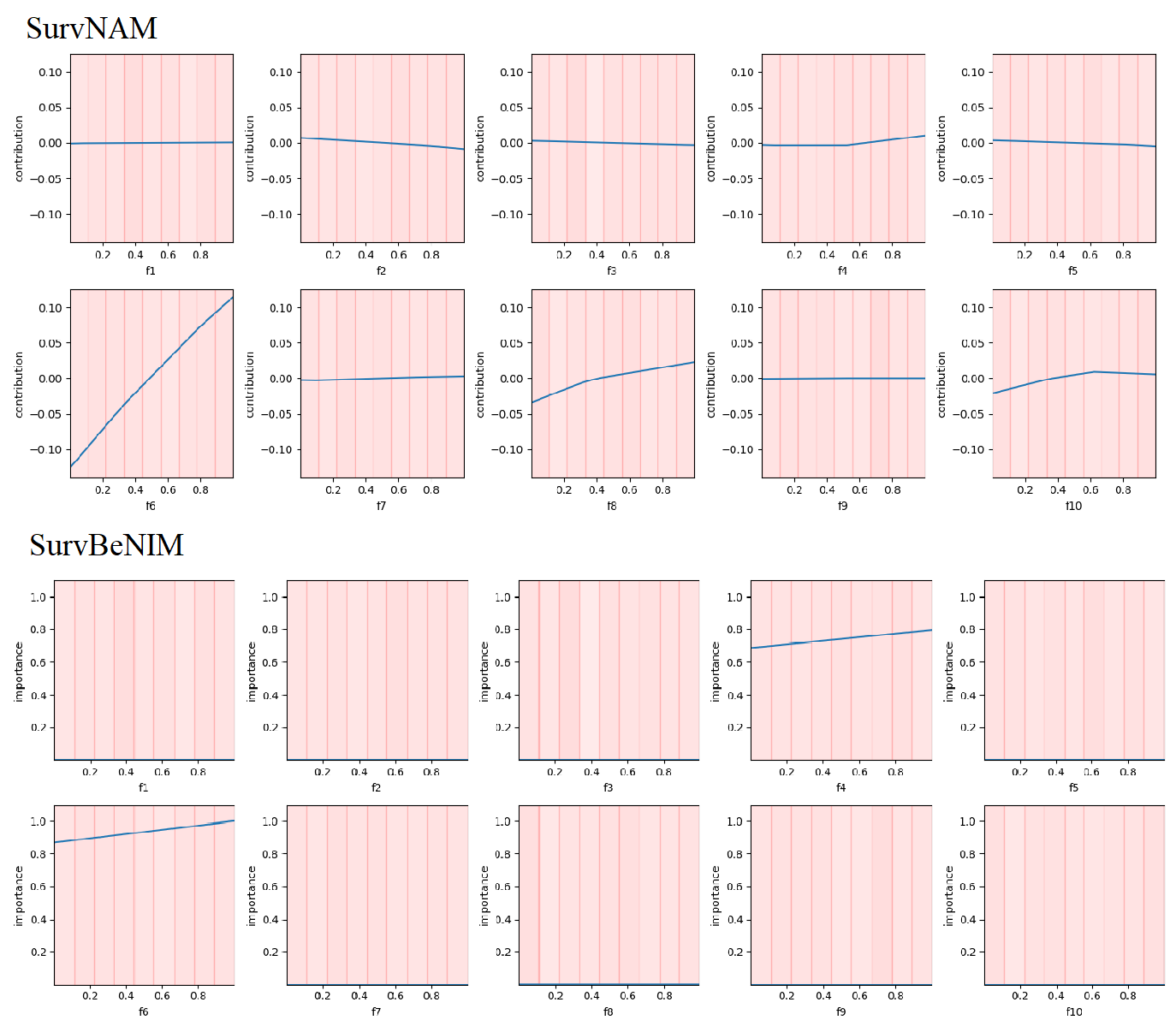}%
\caption{Shape functions obtained by SurvNAM and importance functions obtained
by SurvBeNIM by randomly taking a point from one of five clusters when the Cox
model with $10$ features is used for generating the dataset}%
\label{f:c5f10_shape_imp_122}%
\end{center}
\end{figure}

Shape functions for all $10$ features obtained by means of SurvNAM and
importance functions obtained by means of SurvBeNIM for a randomly selected
point taken from one of $5$ clusters are shown in Figs.
\ref{f:c5f10_shape_imp_122}. The point is from the third cluster which is
generated with vector $\mathbf{b}_{3}^{true}=(0,0,0,0.4,0,0.8,0,0,0,0)$ of the
Cox model. It can be seen from Fig. \ref{f:c5f10_shape_imp_122} that SurvBeNIM
provides correct results demonstrating that the 4-th and the 6-th features are
important. These important features correspond to the vector $\mathbf{b}%
_{3}^{true}$. SurvNAM shows only partly correct results. One can see from
\ref{f:c5f10_shape_imp_122} that SurvNAM selects the 6-th and the 8-th
features as the most important. This incorrect result is caused by the large
number of clusters which violate a structure of the Cox model data.%

\begin{figure}
[ptb]
\begin{center}
\includegraphics[
height=3.2154in,
width=5.4224in
]%
{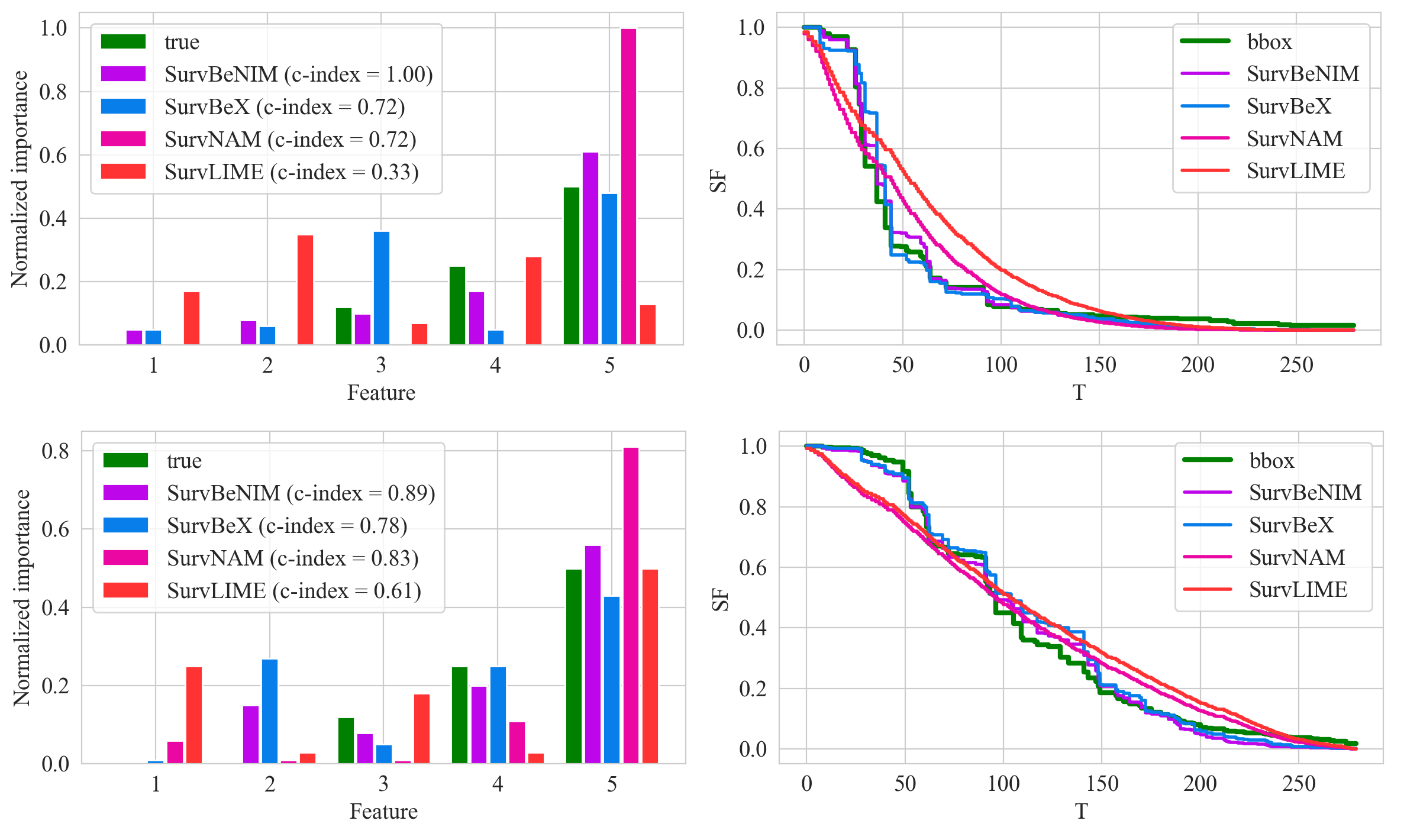}%
\caption{Values of the feature importance (the left column) and SFs provided
by the black-box model (the RSF), SurvLIME, SurvNAM, SurvBeX, and SurvNeNIM
(the right column) for two random instances having 5 features (the first and
second rows)}%
\label{f:c2f5_sf}%
\end{center}
\end{figure}

Fig. \ref{f:c2f5_sf} shows values of the feature importances (the left column)
and SFs provided by the black box, SurvLIME, SurvNAM, SurvBeX, and SurvBeNIM
(the right column) under condition of using 2 clusters and 5 features. Two
instances for explaining are randomly selected from the dataset. It can be
seen from Fig. \ref{f:c2f5_sf} that SFs produced by the black-box model
(depicted by the thicker line) and SurvBeNIM are very close to each other. It
should be noted that SurvBeX is also a good approximation of the SFs whereas
SFs provided by SurvLIME and SurvNAM are far from SFs produced by the
black-box model. One can also see from Fig. \ref{f:c2f5_sf} that the C-index
of SurvBeNIM exceeds C-indices of other models.%

%

\begin{figure}
[ptb]
\begin{center}
\includegraphics[
height=3.1886in,
width=5.4034in
]%
{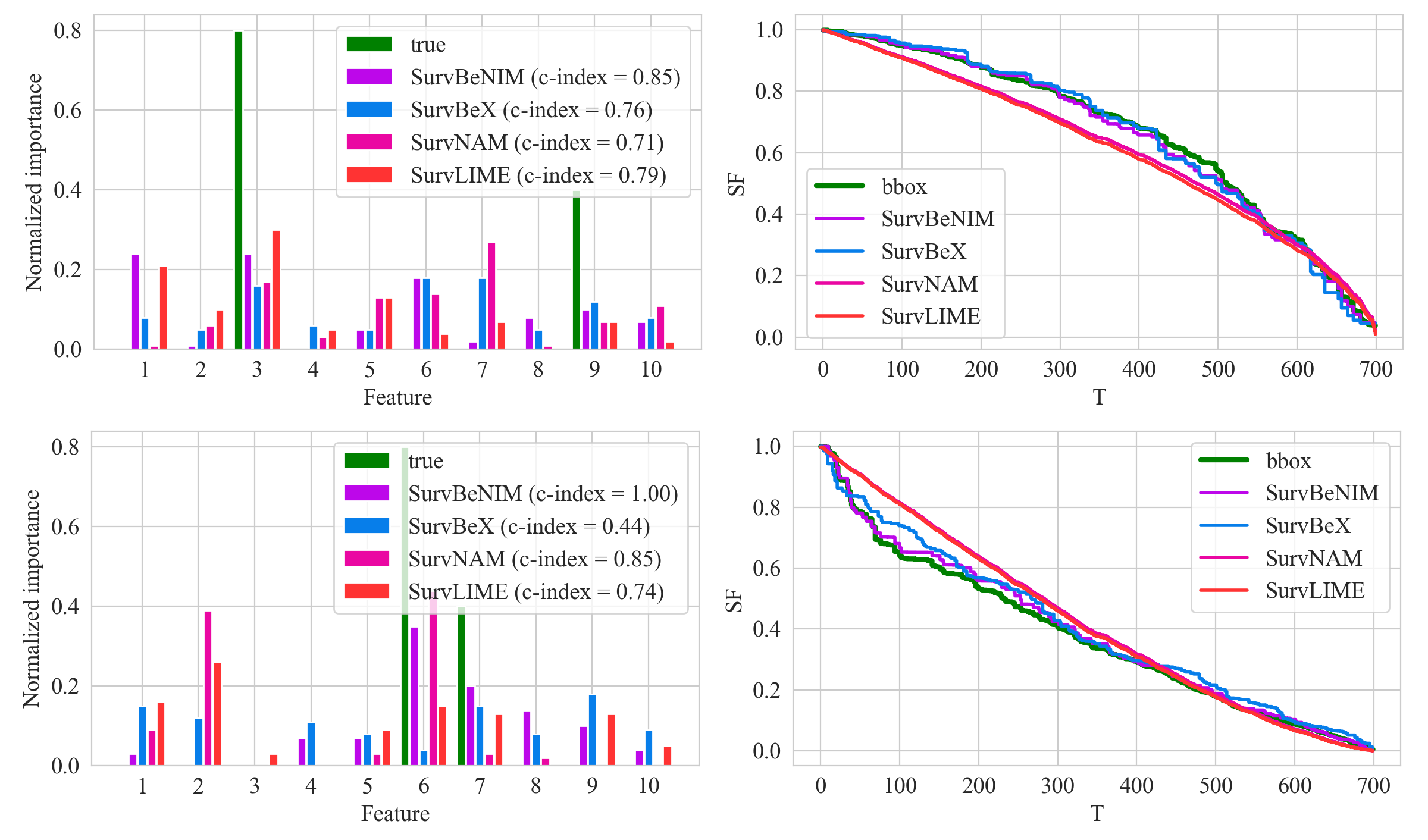}%
\caption{Values of the feature importance (the left column) and SFs provided
by the black-box model (the RSF), SurvLIME, SurvNAM, SurvBeX, and SurvNeNIM
(the right column) for two random instances taken from 5 clusters and having
10 features (the first and second rows)}%
\end{center}
\end{figure}

\begin{figure}
[ptb]
\begin{center}
\includegraphics[
height=3.2802in,
width=5.4656in
]%
{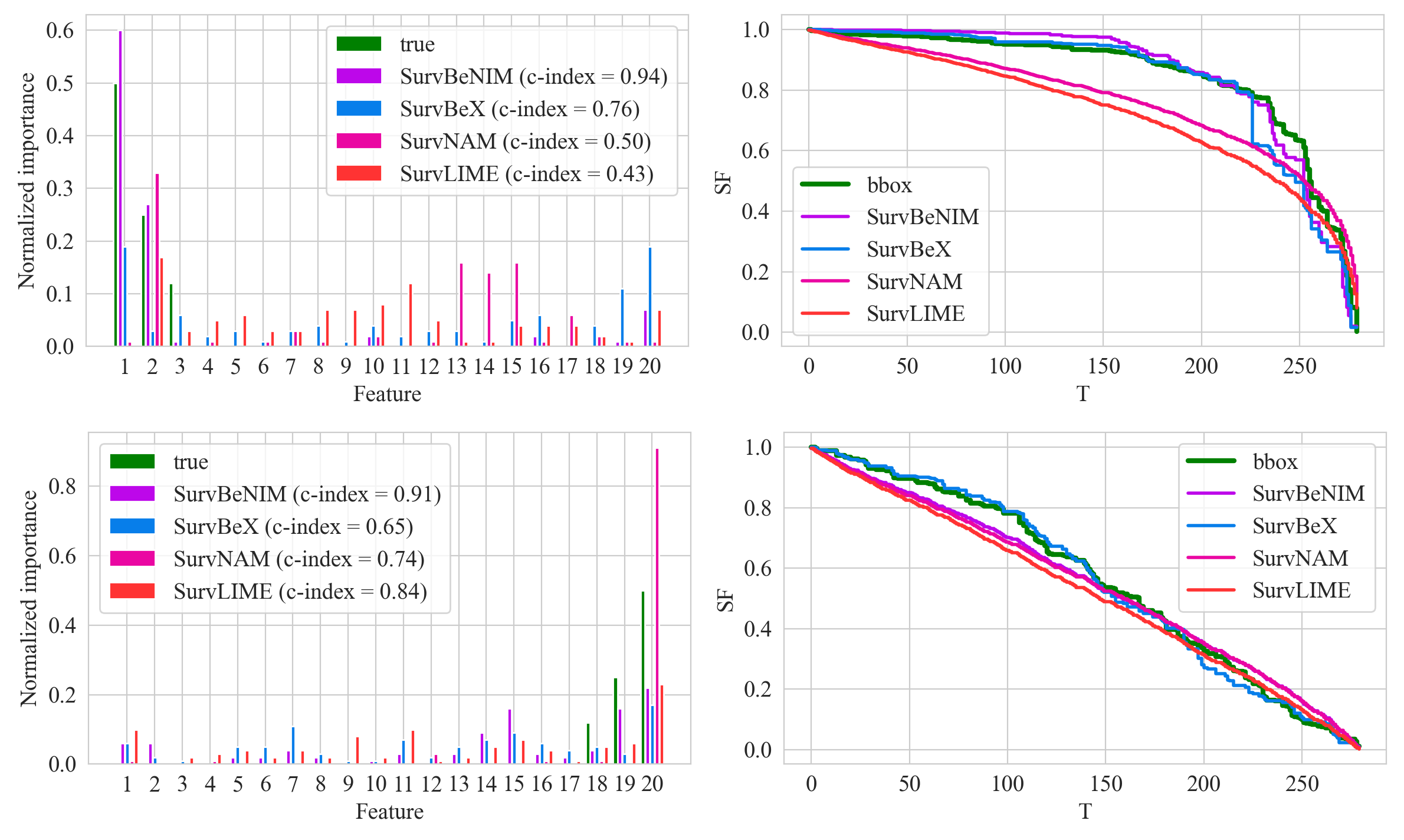}%
\caption{Values of the feature importance (the left column) and SFs provided
by the black-box model (the RSF), SurvLIME, SurvNAM, SurvBeX, and SurvNeNIM
(the right column) for two random instances taken from 2 clusters and
consisting of 20 features (the first and second rows)}%
\label{f:c2f20_sf}%
\end{center}
\end{figure}

The same pictures can be found in Fig. \ref{f:c2f20_sf} where instances with
20 features are considered.%

Table \ref{t:summarize2_5} compares different explanation methods (SurvBeNIM,
SurvBex, SurvNAM, and SurvLIME) when two clusters are considered, the
black-box is the RSF, and the number of features in training instances is 5.
It can be seen from \ref{t:summarize2_5} that SurvBeNIM outperforms other
explanation method. Similar results can be observed in Tables
\ref{t:summarize5_20} and \ref{t:summarize5_10} where the explanation methods
are compared for cases of 2 clusters, 20 features and 5 clusters, 10 features, respectively.%

\begin{table}[tbp] \centering
\caption{Mean values and standard deviations of measures MSD, MKL, and MCI
for SurvBeNIM, SurvBex, SurvNAM, and SurvLIME by 2 clusters and 5 features}%
\begin{tabular}
[c]{cccc}\hline
Method & MSD & MKL & MCI\\ \hline
SurvBeNIM & $0.0127$ & $0.1544$ & $0.8972$\\
SurvBeX & $0.0167$ & $0.2162$ & $0.8182$\\
SurvNAM & $0.0339$ & $0.1243$ & $0.8220$\\
SurvLIME & $0.0298$ & $0.3242$ & $0.5483$\\ \hline
\end{tabular}
\label{t:summarize2_5}%
\end{table}%
%

\begin{table}[tbp] \centering
\caption{Mean values and standard deviations of measures MSD, MKL, and MCI
for SurvBeNIM, SurvBex, SurvNAM, and SurvLIME by 2 clusters and 20 features}%
\begin{tabular}
[c]{cccc}\hline
Method & MSD & MKL & MCI\\ \hline
SurvBeNIM & $0.0094$ & $0.1531$ & $0.7245$\\
SurvBeX & $0.0158$ & $1.0206$ & $0.6481$\\
SurvNAM & $0.0118$ & $0.6977$ & $0.7014$\\
SurvLIME & $0.0124$ & $0.8734$ & $0.6306$\\ \hline
\end{tabular}
\label{t:summarize5_20}%
\end{table}%
%

\begin{table}[tbp] \centering
\caption{Mean values and standard deviations of measures MSD, MKL, and MCI
for SurvBeNIM, SurvBex, SurvNAM, and SurvLIME by 5 clusters and 10 features}%
\begin{tabular}
[c]{cccc}\hline
Method & MSD & MKL & MCI\\ \hline
SurvBeNIM & $0.0223$ & $0.4929$ & $0.8676$\\
SurvBeX & $0.0424$ & $1.1283$ & $0.6188$\\
SurvNAM & $0.0309$ & $0.4453$ & $0.7365$\\
SurvLIME & $0.0258$ & $0.8029$ & $0.7947$\\ \hline
\end{tabular}
\label{t:summarize5_10}%
\end{table}%

\subsubsection{Global explanation: Experiments with synthetic data}

\begin{figure}
[ptb]
\begin{center}
\includegraphics[
height=2.3367in,
width=5.6429in
]%
{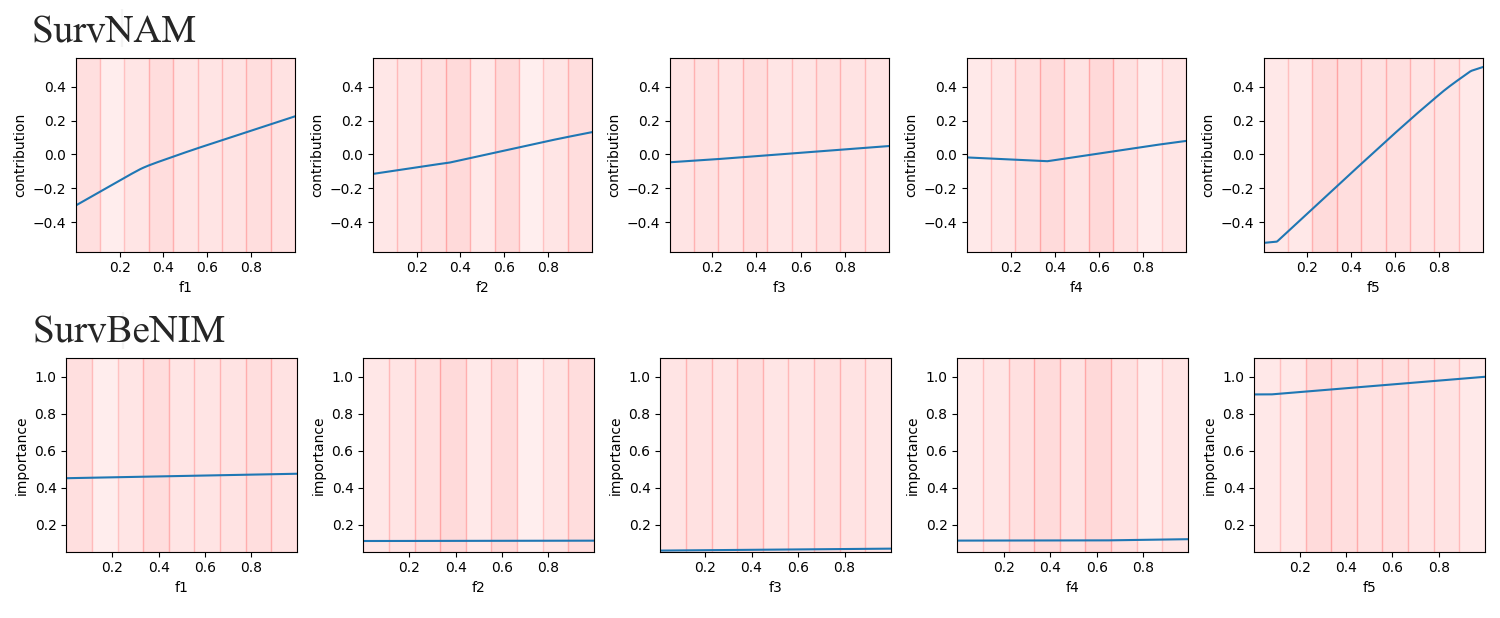}%
\caption{Shape functions obtained by SurvNAM and importance functions obtained
by SurvBeNIM by generating points in accordance with the standard Cox model
with $5$ features}%
\label{f:cox_data_functions}%
\end{center}
\end{figure}

Let us consider the global explanation. The explanation models are learned on
$N=100$ points generated around every testing point from the generated
dataset. First, we study the case when the training set for learning the RSF
is generated in accordance with the standard Cox model. Shape functions for
all features obtained by means of SurvNAM and importance functions obtained by
means of SurvBeNIM are shown in Fig. \ref{f:cox_data_functions}. We can see
that the results provided by SurvNAM and SurvBeNIM are similar, i.e., the
rapidly changing shape functions obtained by using SurvNAM correspond to the
larger values of the importance functions provided by SurvBeNIM. Values of the
MCI for SurvNAM and SurvBeNIM are $0.837$ and $0.907$, respectively.%

In the following experiments, training and testing points are generated in
accordance with the Cox model, but the linear function $\mathbf{b}%
^{true}\mathbf{x}^{\mathrm{T}}$ in (\ref{Cox_generator}) is replaced with the
following non-linear function:%
\begin{equation}
x_{1}^{2}+\max(0,x_{2})+|x_{3}|+10^{-20}\cdot x_{4}+10^{-20}\cdot x_{5}.
\label{Gen_expression}%
\end{equation}

\begin{figure}
[ptb]
\begin{center}
\includegraphics[
height=2.3506in,
width=5.713in
]%
{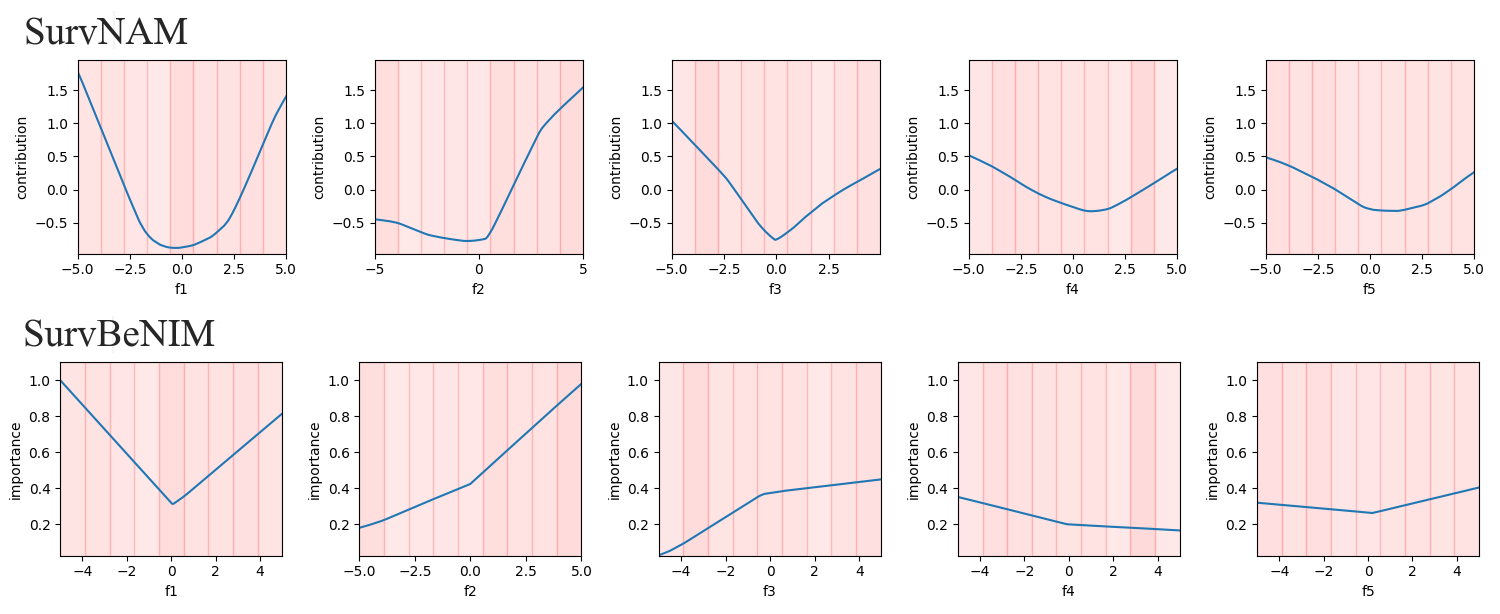}%
\caption{Shape functions obtained by SurvNAM and importance functions obtained
by SurvBeNIM by generating points in accordance with the Cox model having
non-linear relationship between $5$ features}%
\label{f:cox_nonlin_glob}%
\end{center}
\end{figure}

Values of features are generated in accordance with the uniform distribution
with bounds $-5$ and $5$. Shape functions obtained by SurvNAM and importance
functions obtained by SurvBeNIM by generating points in accordance with the
Cox model having non-linear relationship between $5$ features are shown in
Fig. \ref{f:cox_nonlin_glob}. One can see from Fig. \ref{f:cox_nonlin_glob}
that there is again a certain similarity between the shape functions (SurvNAM)
and the importance functions (SurvBeNIM). This can be also seen from the MCI
measure of SurvNAM and SurvBeNIM whose values in this case are $0.901$ and
$0.909$, respectively.%

\begin{figure}
[ptb]
\begin{center}
\includegraphics[
height=2.4033in,
width=5.9214in
]%
{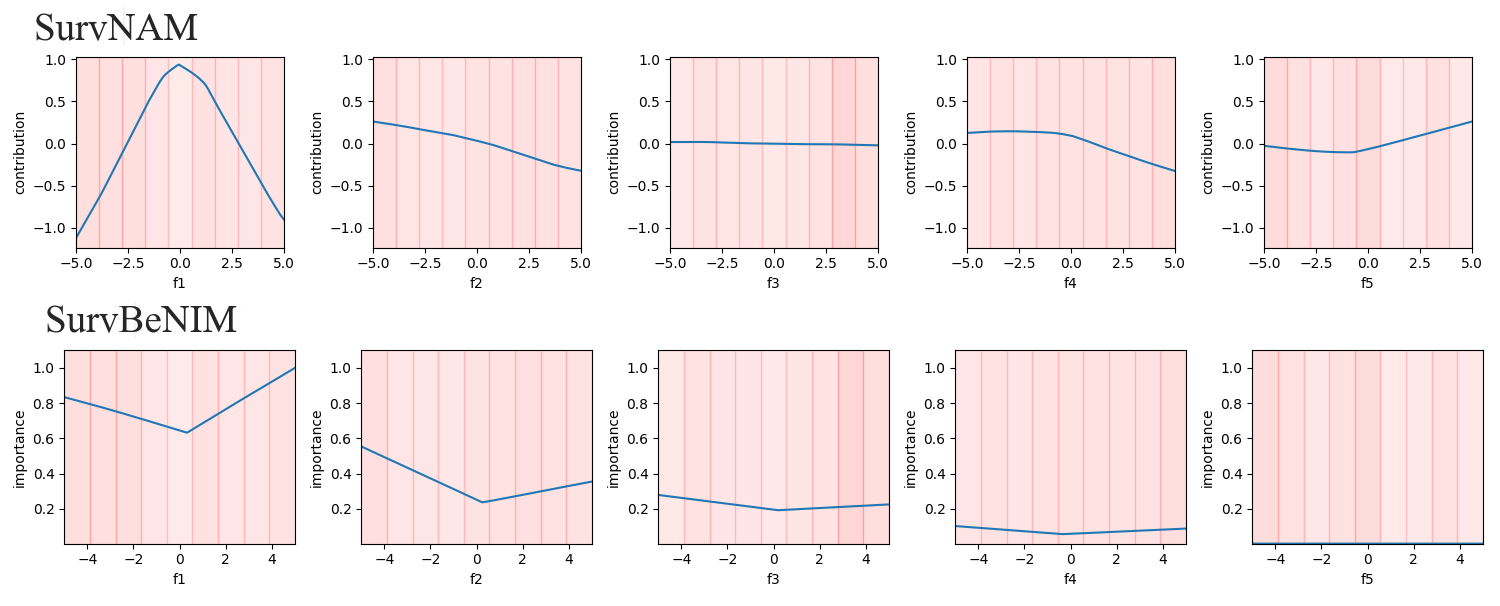}%
\caption{Shape functions obtained by SurvNAM and importance functions obtained
by SurvBeNIM by generating points in accordance with the non-linear function
(\ref{Gen_expression})}%
\label{f:nonlinear_datase}%
\end{center}
\end{figure}

Times to events $T$ for the next experiments are generated by using
(\ref{Gen_expression}) without applying the Cox model, i.e., times to events
are totally determined by (\ref{Gen_expression}) where values of features are
again generated in accordance with the uniform distribution with bounds $-5$
and $5$. The corresponding shape functions obtained by SurvNAM and importance
functions obtained by SurvBeNIM are shown in Fig. \ref{f:nonlinear_datase}.
Values of the MCI for SurvNAM and SurvBeNIM in this case are $0.901$ and
$0.925$, respectively. It can be seen from Fig. \ref{f:nonlinear_datase} that
SurvNAM provides incorrect results for the 3-rd, 4-th and 5-th features. These
results can be explained by the fact that SurvNAM uses the Cox model
approximation, but the generated points are far from generated points in
accordance with the Cox model. The above implies that SurvBeNIM provides
better results when the considered dataset cannot be approximated by the Cox model.%

\subsection{Numerical experiments with real data}

Let us consider numerical experiments with the following real datasets:

\begin{itemize}
\item The \emph{Veterans' Administration Lung Cancer Study (Veteran) Dataset}
\cite{Kalbfleisch-Prentice-1980} contains data on 137 males with advanced
inoperable lung cancer. The subjects were randomly assigned to either a
standard chemotherapy treatment or a test chemotherapy treatment. Several
additional variables were also measured on the subjects. The dataset can be
obtained via the \textquotedblleft survival\textquotedblright \ R package or
the Python \textquotedblleft scikit-survival\textquotedblright \ package..

\item The \emph{German Breast Cancer Study Group 2 (GBSG2) Dataset} contains
observations of 686 women \cite{Sauerbrei-Royston-1999}. Every instance is
characterized by 10 features, including age of the patients in years,
menopausal status, tumor size, tumor grade, number of positive nodes, hormonal
therapy, progesterone receptor, estrogen receptor, recurrence free survival
time, censoring indicator (0 - censored, 1 - event). The dataset can be
obtained via the \textquotedblleft TH.data\textquotedblright \ R package or the
Python \textquotedblleft scikit-survival\textquotedblright \ package.

\item The \emph{Worcester Heart Attack Study (WHAS500) Dataset}
\cite{Hosmer-Lemeshow-May-2008} describes factors associated with acute
myocardial infarction. It considers 500 patients with 14 features. The
endpoint is death, which occurred for 215 patients (43.0\%). The dataset can
be obtained via the \textquotedblleft smoothHR\textquotedblright \ R package or
the Python \textquotedblleft scikit-survival\textquotedblright \ package.
\end{itemize}

\begin{figure}
[ptb]
\begin{center}
\includegraphics[
height=5.0306in,
width=3.806in
]%
{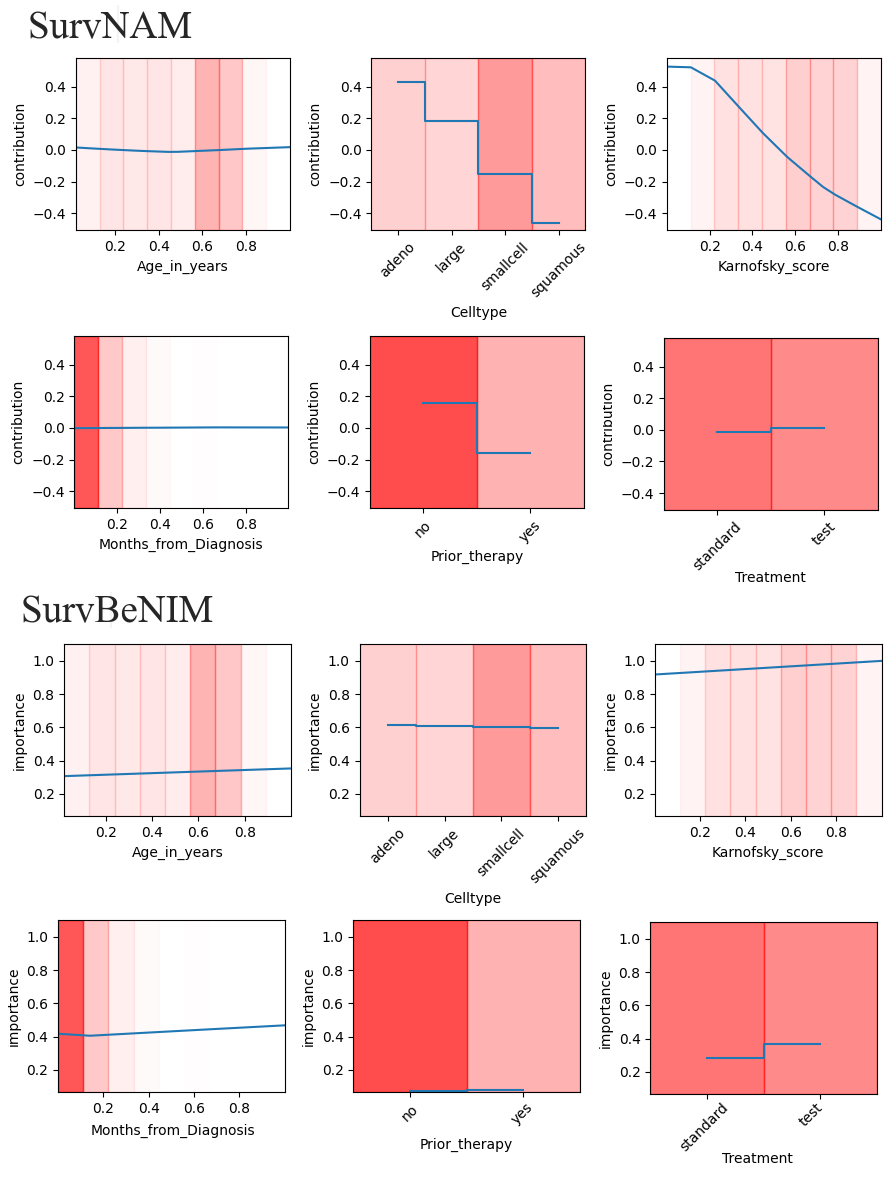}%
\caption{Shape functions obtained by SurvNAM and importance functions obtained
by SurvBeNIM for the Veteran dataset}%
\label{f:veteran_shape_imp}%
\end{center}
\end{figure}

Fig. \ref{f:veteran_shape_imp} shows shape functions obtained by means of
SurvNAM and importance functions obtained by means of SurvBeNIM for the
Veteran dataset. On the one hand, we see that the most important feature
\textquotedblleft Karnofsky score\textquotedblright \ is selected by SurvNAM as
well as SurvBeNIM. Another important feature is \textquotedblleft
Celltype\textquotedblright. It also selected by SurvNAM and SurvBeNIM. On the
other hand, we can observe a difference between results obtained by SurvNAM
and SurvBeNIM. In particular, SurvNAM does not show the importance of
\textquotedblleft Months from diagnosis\textquotedblright \ whereas the
importance function obtained by SurvBeNIM clearly indicates that this feature
impacts on the RSF prediction. Values of the MCI for SurvNAM and SurvBeNIM
obtained for the Veteran dataset are $0.839$ and $0.868$, respectively.%

\begin{figure}
[ptb]
\begin{center}
\includegraphics[
height=4.414in,
width=4.4512in
]%
{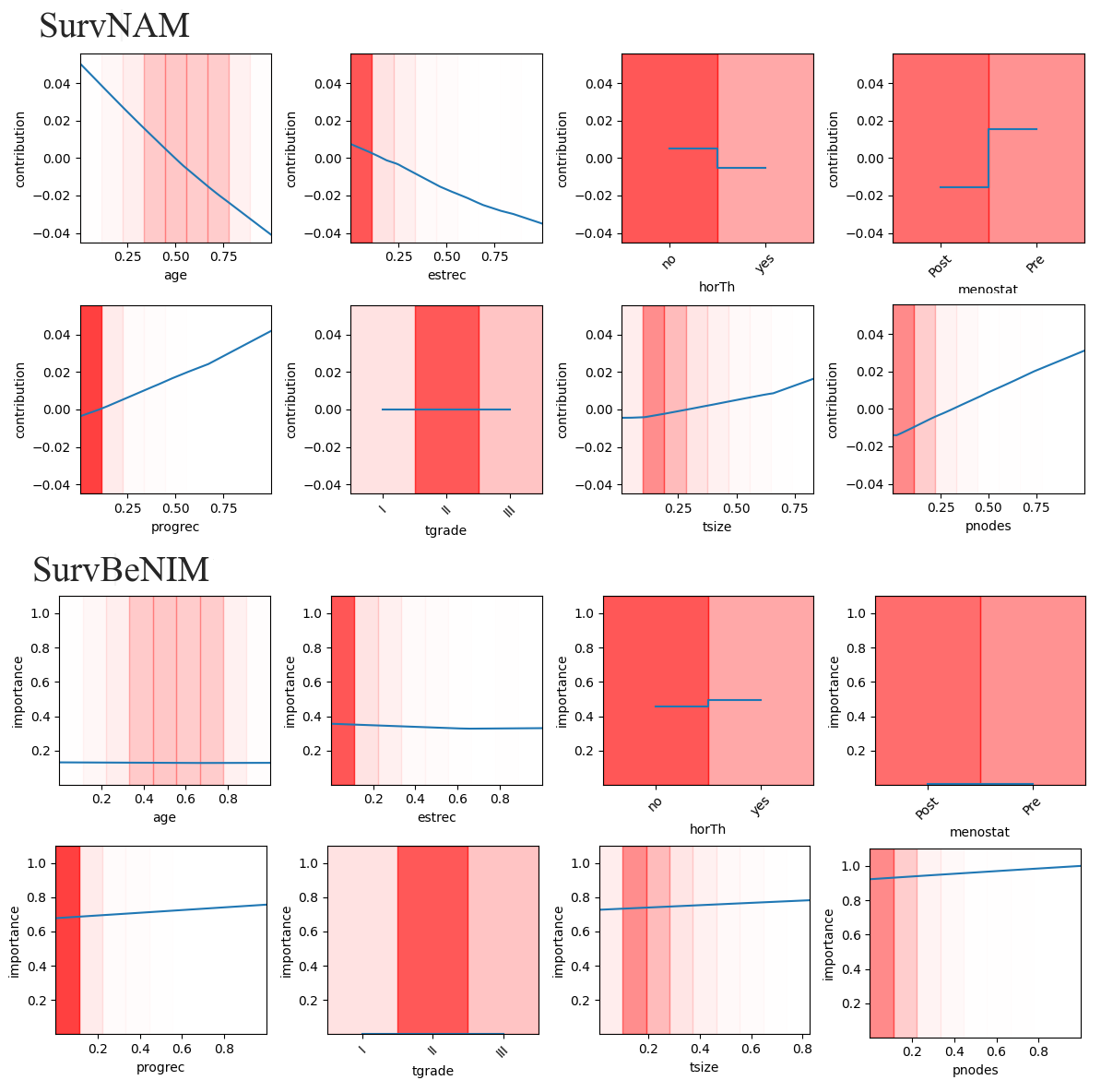}%
\caption{Shape functions obtained by SurvNAM and importance functions obtained
by SurvBeNIM for the GBSG2 dataset}%
\label{f:gbsg_shape_imp}%
\end{center}
\end{figure}

Fig. \ref{f:gbsg_shape_imp} illustrates shape functions provided by SurvNAM
and importance functions provided by SurvBeNIM for the GBSG2 dataset.

One can again see from Fig. \ref{f:gbsg_shape_imp} that the feature
\textquotedblleft pnodes\textquotedblright \ indicating the number of positive
nodes is an important feature in accordance with both the models. The same can
be concluded about features \textquotedblleft progrec\textquotedblright%
\ (progesterone receptor) and \textquotedblleft tsize\textquotedblright%
\ (tumor size). However, SurvNAM\ shows that the feature \textquotedblleft
age\textquotedblright \ is the most important whereas it is not so important in
accordance with SurvBeNIM. It is obvious that SurvNAM provides an incorrect
result by selecting \textquotedblleft age\textquotedblright \ as the most
important feature. Values of the MCI for SurvNAM and SurvBeNIM obtained for
the GBSG2 dataset are $0.627$ and $0.808$, respectively.%

\begin{figure}
[ptb]
\begin{center}
\includegraphics[
height=3.1176in,
width=5.1197in
]%
{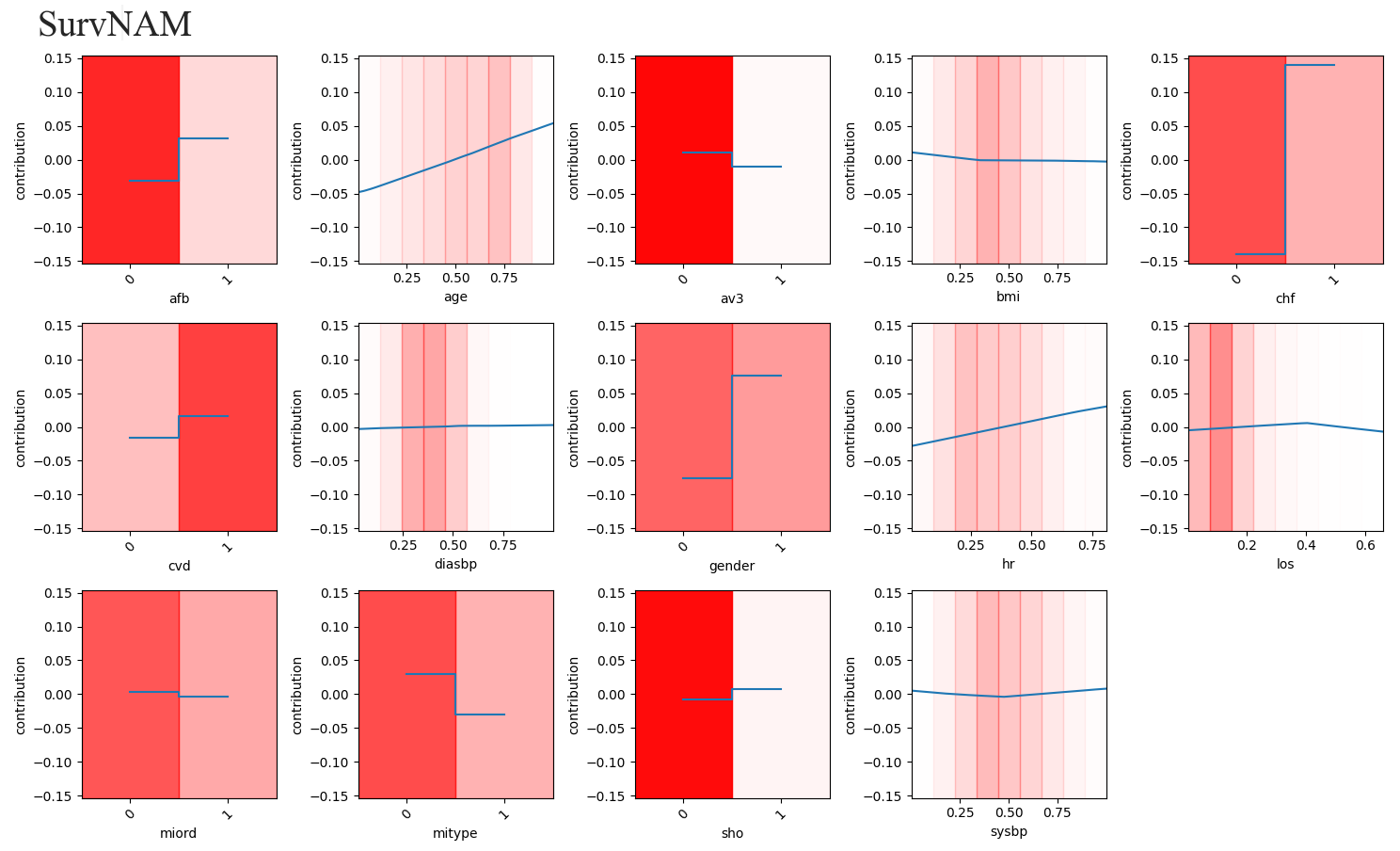}%
\caption{Shape functions obtained by SurvNAM for the WHAS500 dataset}%
\label{f:whas500_shape}%
\end{center}
\end{figure}

\begin{figure}
[ptb]
\begin{center}
\includegraphics[
height=3.1125in,
width=5.1301in
]%
{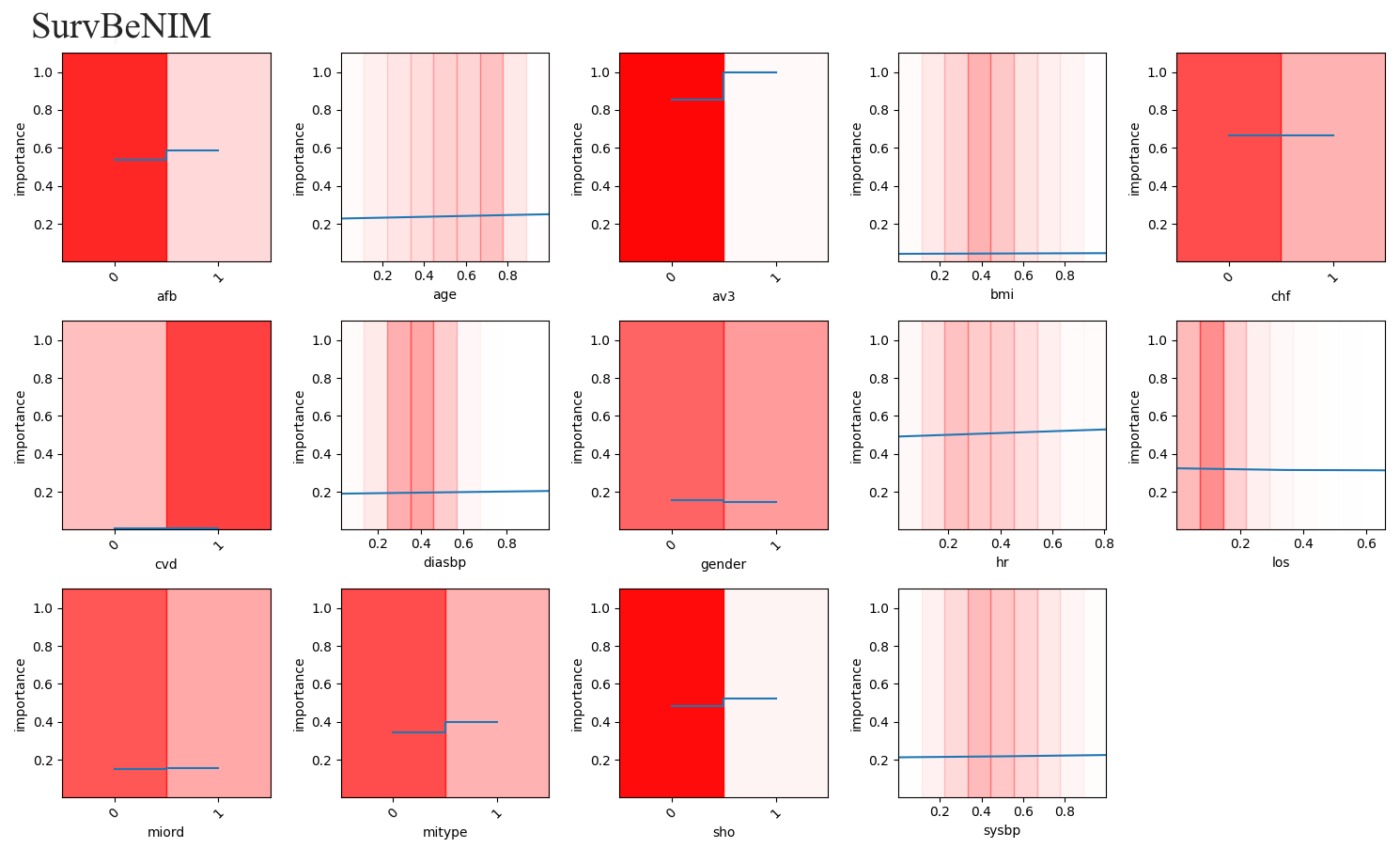}%
\caption{Importance functions obtained by SurvBeNIM for the WHAS500 dataset}%
\label{f:whas500_imp}%
\end{center}
\end{figure}

Shape functions provided by SurvNAM and importance functions provided by
SurvBeNIM for the WHAS500 dataset are depicted in Figs. \ref{f:whas500_shape}
and Fig. \ref{f:whas500_imp}, respectively. We again observe some differences
in selecting the features which impact on the RSF predictions. This difference
is supported by values of the MCI for SurvNAM and SurvBeNIM obtained for the
considered dataset, which are $0.813$ and $0.869$, respectively.%

%

\section{Conclusion}

A method for explaining predictions of machine learning survival models has
been proposed. It can regarded as an extension of methods based on using the
neural additive model (SurvNAM) and the Beran-based explanation method (SurvBeX).

\emph{Advantages} of the proposed method are the following:

\begin{itemize}
\item In contrast to SurvLIME and SurvNAM which use the Cox model for
explanation, the proposed method is based on a more sensitive model, on the
Beran estimator, that takes into account the data structure due to kernels.

\item In contrast to SurvBeX which also uses the Beran estimator, the proposed
method implements the importance functions by means of the neural networks.
The importance functions are more flexible and allow us to observe how the
importance of a feature is changed with values of the feature.

\item In contrast to SurvNAM which provides shape functions which require an
additional interpretation for explanation purposes, SurvBeNIM provides the
importance functions that directly show how features impact on predictions of
the black-box model.

\item Many numerical examples with synthetic and real data have demonstrated
the outperformance of SurvBeNIM in comparison with SurvLIME, SurvNAM, and
SurvBeX. They have also demonstrated that SurvBeNIM provides correct
explanation results for cases when datasets are a complex structure (several
different clusters and complex non-linear relationships between features).
\end{itemize}

The main \emph{disadvantage} of SurvBeNIM is its large computational
difficulty. Due to a large number of hyperparameters of the subnetworks, we
need to train the model implemented SurvBeNIM many times. Moreover, for
correct training the neural networks, we have to generate a larger number of
points around the explained instance due to the large number of training
parameters (the neural network weights). This peculiarity also complicates the
explanation problem.

It should be noted that we have used only the Gaussian kernels for
implementing the Beran estimator. Therefore, an interesting direction for
further research is to study different types of kernels. Moreover, it is
interesting to add training parameters to kernels. This extension of kernels
may provide more flexible and outperforming results. Another direction for
research is to combine SurvNAM and SurvBeNIM. Since both the methods have a
common elements in the form of neural networks, then the corresponding loss
function can consist of two parts: the first part is the loss from SurvNAM;
the second part is the loss from SurvBeNIM. As a result, two functions (the
shape and importance functions) can be simultaneously computed and compared.

\bibliographystyle{unsrt}
\bibliography{Boosting,Classif_bib,Deep_Forest,Explain,Explain_med,IntervalClass,MYBIB,MYUSE,Survival_analysis}

\end{document}